\RequirePackage{fix-cm}
\documentclass[twocolumn,english]{svjour3}

\smartqed
\usepackage{graphicx}
\usepackage{mathptmx}

\usepackage{xcolor}

\usepackage{booktabs}
\usepackage{multicol}
\usepackage{multirow}
\usepackage{microtype}

\usepackage{amsmath}
\usepackage{amssymb}
\usepackage{authblk}
\usepackage{algorithm}
\usepackage{algorithmicx}
\usepackage{algpseudocode}

\usepackage{bm}
\usepackage{url}
\usepackage[hidelinks]{hyperref}
\usepackage[sort&compress,numbers]{natbib}

\usepackage{pifont}
\newcommand{\cmark}{\ding{51}}
\newcommand{\xmark}{\ding{55}}

\journalname{Data Mining and Knowledge Discovery}

\begin{document}

\title{TCMI: a non-parametric mutual-dependence estimator for multivariate continuous distributions}
\titlerunning{TCMI: a non-parametric mutual-dependence estimator}
\author{B. Regler$^1$ \and M. Scheffler$^{1,2}$ \and L. M. Ghiringhelli$^{1,2}$}
\institute{
    ${}^1$ The NOMAD Laboratory at the Fritz Haber Institute of the Max Planck Society and Humboldt University, Berlin, Germany\\
    ${}^2$ Physics Department and IRIS Adlershof, Humboldt-Universität zu Berlin, Berlin, Germany\\[1em]
    \email{regler@fhi-berlin.mpg.de}
}
\date{Received: date / Accepted: date}
\maketitle

\begin{abstract}
    The identification of relevant features, i.e., the driving variables that determine a process or the properties of a system, is an essential part of the analysis of data sets with a large number of variables. A mathematical rigorous approach to quantifying the relevance of these features is mutual information. Mutual information determines the relevance of features in terms of their joint mutual dependence to the property of interest. However, mutual information requires as input probability distributions, which cannot be reliably estimated from continuous distributions such as physical quantities like lengths or energies. Here, we introduce total cumulative mutual information (TCMI), a measure of the relevance of mutual dependences that extends mutual information to random variables of continuous distribution based on cumulative probability distributions. TCMI is a non-parametric, robust, and deterministic measure that facilitates comparisons and rankings between feature sets with different cardinality. The ranking induced by TCMI allows for feature selection, i.e., the identification of variable sets that are nonlinear statistically related to a property of interest, taking into account the number of data samples as well as the cardinality of the set of variables. We evaluate the performance of our measure with simulated data, compare its performance with similar multivariate-dependence measures, and demonstrate the effectiveness of our feature-selection method on a set of standard data sets and a typical scenario in materials science.

\end{abstract}

\section{Introduction}
\label{sec:introduction}

The past two decades have been marked by an explosion in the availability of scientific data and significant improvements in statistical data analysis. In particular, the physical sciences have seen an unprecedented surge in data exploration aimed at the data-driven discovery of statistical dependencies of physical variables relevant to a property of interest. These observations culminated in the emergence of a new paradigm in science, the so-called ``big-data driven'' science \citep{HeyTansleyEtAl:2009}.

\begin{figure}[t]
    \centering
    \includegraphics[width=\columnwidth]{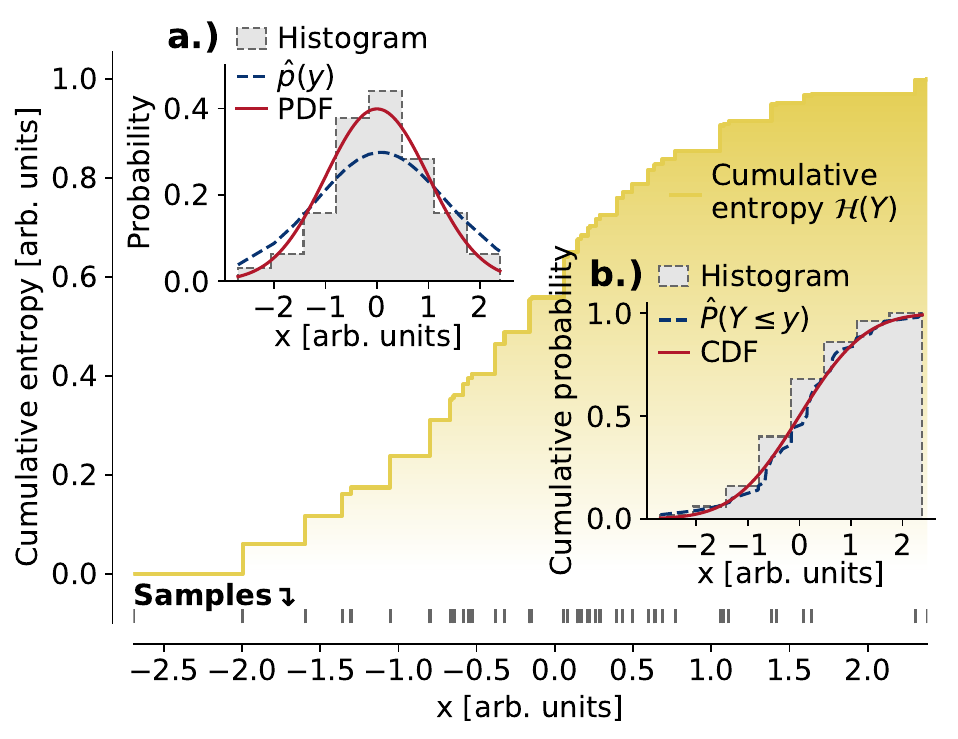}
    \caption{Empirical cumulative entropy $\hat{\mathcal H}(Y)$ of a normal distribution for $50$ data samples, which are shown as ticks in the bottom of the figure. Insets a.) and b.) show the (ground-truth) probability density (PDF) and cumulative probability (CDF) of the normal distribution, empirical cumulative distribution, $\hat{P}(Y \leq y)$, and estimated probability density, $\hat{p}(y)$. The estimated probability density was obtained by optimizing the bandwidth of a kernel-density estimator through 10-fold cross-validation. Further, histograms of PDF and CDF are also drawn to provide an example of how continuous distributions can be approximated by discrete discontinuous functions.}
    \label{fig:cumulative_entropy_visualization}
\end{figure}

\subsection{Feature selection}

The identification of relevant variables, i.e., the properties or the driving variables of a process or system's property, has propelled investigations for an understanding of the underlying processes that generated the data \citep{GuyonElisseeff:2003}. Such a variable $X \in \vec{X}$ may be an attribute, parameter, or a combination of properties measured or obtained from experiments or simulations. The fundamental challenge is to find a functional dependency $f: \vec{X} \mapsto Y$, between a set of variables $\vec{X}' \in \vec{X}$ related to a certain output $Y$ (target, response function). The objective is to find a set of variables (the so-called features) that maximizes a feature-selection criterion $\mathcal Q$ with respect to a property of interest $Y$ \citep{BlumLangley:1997,KohaviJohn:1997},
\begin{equation}
    \vec{X}^* = \operatorname*{arg\,max}_{\vec{X}' \subseteq \vec{X}} \mathcal Q(Y; \vec{X}') \ .
    \label{eq:feature_selection}
\end{equation}
Feature selection comprises two parts: (i) the choice of a search strategy and (ii) a feature-selection criterion $\mathcal Q$ for evaluating a feature-subset's relevance.

\subsubsection*{Search strategies}

There are several search strategies to identify the relevant features of a data set \citep{NarendraFukunaga:1977,SiedleckiSklansky:1993,PudilNovovicovaEtAl:1994,EberhartKennedy:1995,MichalewiczFogel:2004}, ranging from optimal solvers (such as exhaustive search or accelerated methods based on the monotonic property of a feature-selection criterion), to sub-optimal solvers (such as greedy, heuristic, or stochastic solvers) \citep{GuyonElisseeff:2003,KohaviJohn:1997,NarendraFukunaga:1977,SiedleckiSklansky:1993,PudilNovovicovaEtAl:1994,Whitney:1971,PudilNovovicovaEtAl:2002,MarillGreen:1963,LandDoig:1960,YuYuan:1993,Clausen:1999,MorrisonJacobsonEtAl:2016,ForsatiMoayedikiaEtAl:2011,Reunanen:2006}. Optimal solvers explore all feature-subset combinations for a global optimum and, as such, are generally impractical for data sets with a large number of features due to cost and time constraints on computer resources. Sub-optimal search strategies (e.g., sequential floating forward selection \citep{PudilNovovicovaEtAl:1994,Whitney:1971}, sequential backward elimination \citep{MarillGreen:1963}, and minimal-redundancy-maximal-relevance criterion \citep{PengLongEtAl:2005}), conversely, balance accuracy and speed, but may not find the optimal set of features with respect to a targeted property. A search strategy that can be used both as an optimal or sub-optimal solver, is branch and bound \citep{NarendraFukunaga:1977,PudilNovovicovaEtAl:2002,LandDoig:1960,YuYuan:1993,Clausen:1999,MorrisonJacobsonEtAl:2016}. Branch and bound implicitly performs an exhaustive search, but uses an additional bounding criterion to discard feature subsets, whose feature-selection criteria are lower than the feature-selection criterion of the current best feature subset in the search.

\subsubsection*{Feature-selection criterion}

The feature-selection criterion $\mathcal Q$ can be used as a score that allows the identified features to be ranked by relevance prior to subsequent data analyses. The academic community has extensively explored several feature-selection criteria to evaluate a feature's relevance \citep{KhaireDhanalakshmi:2019}, including distance measures \citep{Basseville:1989,AlmuallimDietterich:1994}, dependency measures \citep{Modrzejewski:1993}, consistency measures \citep{Arauzo-AzofraBenitezEtAl:2008}, and information measures \citep{VergaraEstevez:2014}. Ideally, feature-selection criteria are not restricted to specific type of dependencies, are robust against imprecise values in the data, and are deterministic, i.e., such that the feature selection is consistent and reproducible for the same set of variables, type of settings, and data. The prevailing method for quantifying multivariate dependences is mutual information, which determines the relevance of variables in terms of their joint mutual dependence to a property of interest \citep{Shannon:1948}.

There are several reasons to consider mutual-information-based quantities for feature selection. The two most important reasons are: (i) mutual information quantifies multivariate nonlinear statistical dependencies and (ii) mutual information provides an intuitive quantification of the relevance for a feature subset $\vec{X}' \subseteq \vec{X}$ relative to an output $Y$ \citep{VergaraEstevez:2014}: it is bounded from below (for statistically independent variables with respect to an output) and can be bounded from above (for functional dependence variables), increases as the number of sample sizes increases, and quantifies the strength of the dependence based on a mathematical rigorous framework from communication theory \citep{Shannon:1948}. However, mutual information requires probability distributions, which are problematic for high-dimensional data sets and are difficult to obtain from real-valued data samples of continuous distributions.

\begin{table*}[t]
    \centering
    \begin{tabular}{llc}
        \toprule
        \textbf{Abbreviation} & \textbf{Explanation} & \textbf{Reference} \\
        \midrule
        CMI & Cumulative mutual information & \cite{NguyenMuellerEtAl:2013} \\
        MAC & Multivariate maximal correlation analysis & \cite{NguyenMuellerEtAl:2014a} \\
        UDS & Universal dependency analysis & \cite{NguyenMandrosEtAl:2016,WangRomanoEtAl:2017} \\
        MCDE & Monte Carlo dependency estimation & \cite{FoucheBoehm:2019,FoucheMazankiewiczEtAl:2021}\\
        TCMI & Total cumulative mutual information & This work \\
        \bottomrule
    \end{tabular}
    \caption{Abbreviations used in the manuscript.}
    \label{tab:abbreviations}
\end{table*}

\subsection{Our approach}

We propose total cumulative mutual information (TCMI): a non-parametric, robust, and deterministic measure of the relevance of mutual dependences of continuous distributions between variable sets of different cardinality. TCMI can be applied if the dependence between a set of variables and an output is not yet known and the dependence is nonlinear and multivariate. Like mutual information, TCMI relates the strength of the dependence between a set of variables and an output to the number of data samples. In addition, TCMI relates the strength of the dependence to the cardinality of the subsets. Thus, TCMI allows an unbiased comparison between different sets of variables without depending on externally adjustable parameters.

TCMI is based on cumulative mutual information and inherits many of the properties of mutual-information based feature-selection measures: it is bounded from below and above and monotonically increases the more features are subsequently added to a candidate feature set, but only until all variables related to an output are included. In contrast to other feature-selection measures based on cumulative mutual information, TCMI uses cumulative probability distributions. Cumulative probability distributions can be directly obtained from empirical data of continuous distributions, without the need to quantize the set of variables prior to estimating a feature subset's dependence to a property of interest.

We combine TCMI with the branch-and-bound algorithm \citep{NarendraFukunaga:1977,PudilNovovicovaEtAl:2002,LandDoig:1960,YuYuan:1993,Clausen:1999,MorrisonJacobsonEtAl:2016}, which has proven to be efficient in the discovery of nonlinear functional dependencies \citep{ZhengKwoh:2011,MandrosBoleyEtAl:2017}. TCMI therefore identifies a set of variables that are statistically related to an output. As TCMI is model independent, a functional relationship must be constructed (in the following referred to as a model) to relate these features with an output. The model construction is not part of this work, but can be done, for example, through data-analytics techniques such as symbolic regression, both in the genetic-programming \citep{Koza:1994} and in compressed-sensing implementations \citep{GhiringhelliVybiralEtAl:2015,OuyangCurtaroloEtAl:2018}, or regression tree-based approaches \citep{BreimanFriedmanEtAl:1984}.

In brief, our feature-selection procedure can be divided into three steps: In the first step, we quantify the dependence between the set of features and an output as the difference between cumulative marginal and cumulative conditional distributions. In the second step, we estimate the relevance of a feature set by comparing its strength of dependence to the mean dependence of features under the assumption of independent random variables. In the third step, we identify a set of relevant features with the branch-and-bound algorithm to find the set of variables from an initial list that best characterizes an output.

\begin{table*}[t]
    \centering
    \begin{tabular}{ll}
        \toprule
        \textbf{Symbol} & \textbf{Definition} \\
        \midrule

        $Y$ & Output, target, response function \\
        $X$, $\vec{X}$, $X'$, $\vec{X}'$ & Features, variables \\
        $D_\text{KL}(U \| V)$ & Kullback-Leibler divergence of two distributions $U$ and $V$ \\[1em]

        \multicolumn{2}{@{} l}{\itshape Discrete data} \\
        $p(y)$ & (Marginal) probability density of $y \in Y$ \\
        $p(x, y)$ & Joint probability density of $x \in X$ and $y \in Y$ \\
        $p(y \vert x)$ & Conditional probability density of $y \in Y$ given $x \in X$ \\
        $H(Y)$ & Shannon entropy of $Y$ \\
        $I(Y; X)$ & Shannon mutual information \\
        $I_0(Y; X)$ & Baseline correction term of the Shannon mutual information \\
        $D(Y; X)$ & Normalized mutual information/ fraction of information \\[1em]

        \multicolumn{2}{@{} l}{\itshape Continuous data} \\
        $P(y)$, $P(Y \leq y)$ & (Marginal) cumulative distribution of $y \in Y$ \\
        $P'(x)$, $P'(X \leq x)$ & (Marginal) residual cumulative distribution of $x \in X$ \\
        $P(x, y)$, $P(X \leq x, Y \leq y)$ & Joint cumulative distribution of $x \in X$ and $y \in Y$ \\
        $P(y \vert x)$, $P(Y \leq Y \vert X \leq x)$ & Conditional cumulative distribution of $y \in Y$ given $x \in X$ \\
        $\mathcal H(Y)$ & Cumulative entropy of $Y$ \\
        $\mathcal I(Y; X)$, $\mathcal I^*(Y; X)$ & (Adjusted) cumulative mutual information \\
        $\mathcal I_0(Y; X)$ & Baseline correction term for cumulative mutual information \\
        $\mathcal D(Y; X)$, $\mathcal D^*(Y; X)$ & (Adjusted) fraction of cumulative information \\[1em]

        \multicolumn{2}{@{} l}{\itshape Empirical estimates} \\
        $\mathcal E$ & Estimator \\
        $\hat{\mathcal E}$ & Empirical estimator, e.g., $\hat{\mathcal H}(Y)$ \\
        $\mathcal E_0$ & Baseline adjustment term of an estimator $\mathcal E$ \\[1em]

        \multicolumn{2}{@{} l}{\itshape Total cumulative mutual information (TCMI)} \\
        $\hat{\mathcal D}_\text{min}^*(Y; X)$ & Minimum adjusted contribution of empirical cumulative mutual information \\
        $\langle \hat{\mathcal D}_\text{TCMI}^*(Y; X) \rangle$ & Average adjusted contribution of cumulative mutual information (TCMI) \\[1em]

        \multicolumn{2}{@{} l}{\itshape Branch and bound} \\
        $\bar{\mathcal Q}$ & Bounding criterion \\
        $\mathcal{Q}$, $\mathcal{Q}^*$ & (Adjusted) feature-selection criterion \\
        \bottomrule
    \end{tabular}
    \caption{List of symbols and notations used in this paper.}
    \label{tab:notation}
\end{table*}

\subsection{Outline}

The remainder of this work is organized as follows. Section~\ref{sec:related_work} discusses the relationship between TCMI and previous work. Section~\ref{sec:theoretical_background} introduces the theoretical background of cumulative mutual information. Section~\ref{sec:empirical_estimations_of_cumulative_entropy_and_cumulative_mutual_information} describes the empirical estimation of cumulative mutual information for continuous distributions from limited sample data. Section~\ref{sec:baseline_adjustment} explains the steps introduced to adjust the cumulative mutual information with respect to the number of data samples and the cardinality of the feature subset. Section~\ref{sec:total_cumulative_mutual_information} introduces TCMI. Section~\ref{sec:feature_selection} describes the implementation details of the feature-subset search using the branch-and-bound algorithm in detail. Section~\ref{sec:experiments} reports on the performance evaluation of TCMI on generated data, standard data sets, and on a typical scenario in materials science. In the same section, TCMI is also compared with similar multivariate dependence measures such as cumulative mutual information (CMI: \cite{NguyenMuellerEtAl:2013}), multivariate maximal correlation analysis (MAC: \cite{NguyenMuellerEtAl:2014a}), and universal dependency analysis (UDS,  \cite{NguyenMandrosEtAl:2016,WangRomanoEtAl:2017}). We would like to point out that feature selection is a broad area of research and can be achieved using a variety of techniques; this work therefore focuses on feature-selection methods based on mutual information that can be applied prior to subsequent data-analysis tasks. To illustrate this, we provide a real-world example in Section~\ref{sec:octet_binary_compound_semiconductors} by building a model from the identified features using TCMI, and comparing its performance to in-built feature-selection methods that perform feature selection during model construction. Finally, Sections~\ref{sec:discussion} and \ref{sec:conclusions} present the discussion and conclusions of this work. Abbreviations, notations, and terminologies are summarized in Tables~\ref{tab:abbreviations} and \ref{tab:notation}.

\section{Related work}
\label{sec:related_work}

Many dependence measures such as Pearson $R$ and Spearman's rank $\rho$ correlation coefficients \citep{Pearson:1896,Spearman:1904}, distance correlation (DCOR: \cite{SzekelyRizzoEtAl:2007,SzekelyRizzo:2014}), kernel density estimation (KDE: \cite{Scott:1982,Silverman:1986}), or $k$-nearest neighbor estimation ($k$-NN: \cite{KozachenkoLeonenko:1987}) are limited to bivariate dependencies only (Pearson, Spearman), are limited to specific types of dependencies (Spearman, DCOR), or require assumptions about the functional form of $f$ (KDE, $k$-NN). Bivariate extensions \citep{SchmidSchmidt:2007}, KDE, and $k$-NN are further not applicable to high-dimensional data sets.

For high-dimensional data sets, several authors proposed subspace-slicing techniques \citep{FoucheBoehm:2019,FoucheMazankiewiczEtAl:2021,KellerMullerEtAl:2012}, which repeatedly apply conditions on each variable and perform statistical hypothesis tests to estimate the degree of dependence to an output. However, with these methods, all possible combinations of variables must be enumerated and, therefore, are computationally intractable for feature-selection tasks. In addition, the strength of their dependences are not related to the cardinality of feature subsets and therefore cannot be used to compare different sets of variables.

Model-dependent methods such as data-analytics techniques \citep{Koza:1994,GhiringhelliVybiralEtAl:2015,OuyangCurtaroloEtAl:2018,BreimanFriedmanEtAl:1984} perform feature selection while creating a model. Alternatively, \emph{post-hoc} analysis tools such as unified dependence measures \citep{LundbergLee:2017} can be used to assign each variable an importance value for a particular estimation. However, these methods add an additional degree of complexity, which makes it difficult to reliably assess the dependence among variables. Another approach are information-theoretic dependence measures. These measures are based on mutual information and ascertain whether or not the values of a set of variables are related to an output. As a result, they provide a model-independent approach to estimating the strength of dependences between variables.

Multivariate extensions to mutual information (e.g., interaction information \citep{McGill:1954} and total correlation \citep{Watanabe:1960}) require knowledge about the underlying probability distributions and are therefore difficult to estimate: estimations either require large amount of data and needs to be specified for each new data set at hand \citep{BelghaziBaratinEtAl:2018} or are affected by the curse of dimensionality \citep{Bellman:1957} such as the Kozachenko-Leonenko estimator \citep{KozachenkoLeonenko:1987,KraskovStoegbauerEtAl:2004}. Recently, several authors proposed related approaches to extending mutual information: cumulative mutual information (CMI: \cite{NguyenMuellerEtAl:2013}), multivariate maximal correlation analysis (MAC: \cite{NguyenMuellerEtAl:2014a}), and universal dependency analysis (UDS: \cite{NguyenMandrosEtAl:2016,WangRomanoEtAl:2017}). All three methods estimate the strength of dependence between a set of variables from cumulative probability distributions, and thus can be viewed as alternative measures of uncertainty that extend Shannon entropy (and mutual information) to random variables of multivariate continuous distributions.

CMI quantifies multivariate mutual dependences by considering the cumulative distributions of variables and by heuristically approximating the conditional cumulative entropy via data summarization and clustering. MAC is based on Shannon entropy over discretized data which MAC obtains by maximizing the normalized total correlation with respect to cumulative entropy. UDS uses optimal discretization to compute the conditional cumulative entropy, where the Shannon entropy defines the number of bins required to estimate the conditional cumulative entropy. Since all three dependence measures expose adjustable parameters to optimize the quantization of continuous distributions, the choice of these parameters therefore has a strong impact on the strength of mutual dependence between a set of variables $\vec{X} = \{ X_1, \ldots, X_n \}$ and an output $Y$, and thus on the ranking induced by the relevance function and a feature-selection criterion. For feature selection, these measures are impractical.

Our approach, TCMI, extends CMI, but does not require to quantize real-valued data samples of continuous data distributions to estimate the joint cumulative distribution of continuous distributions. TCMI therefore does not require data summarization techniques to estimate multivariate dependences between continuous distributions, unlike similar approaches such as MAC and UDS. TCMI is non-parametric, as opposed to other estimation methods based on mutual information, e.g., neural networks \citep{BelghaziBaratinEtAl:2018} or mutual-information-based feature selection algorithms originally developed for discrete data \citep{KwakChong-Ho:2002,ChowHuang:2005,EstevezTesmerEtAl:2009,HuZhangEtAl:2011,ReshefReshefEtAl:2011,BennasarHicksEtAl:2015}. TCMI therefore allows to reliably compare the strength of dependence between different sets of variables. On top of that, TCMI relates the strength of a dependence to a feature-subset's cardinality and the number of data samples by basing the score on the dependence of the same set of variables under the independence assumption of random variables \citep{VinhEppsEtAl:2009,VinhEppsEtAl:2010}.

\section{Theoretical background}
\label{sec:theoretical_background}

Mutual information and all measures presented in the following quantify relevance by means of the similarity between two distributions $U(\vec{X}, Y)$ and $V(\vec{X}, Y)$ using Kullback-Leibler divergence, $D_\text{KL}(U(\vec{X}, Y) \| V(\vec{X}, Y))$ \citep{KullbackLeibler:1951}. They do not require any explicit modeling to quantify linear and nonlinear dependencies, monotonically increase with the cardinality of a feature's subset $\vec{X}' \subseteq \vec{X}$,
\begin{multline}
    \min_{X \in \vec{X}'}  D_\text{KL}(U(\vec{X}' \setminus X, Y) \| V(\vec{X}' \setminus X, Y)) \\
    \leq D_\text{KL}(U(\vec{X}', Y) \| V(\vec{X}', Y)) \ ,
    \label{eq:monotonicity_information_theory}
\end{multline}
and are invariant under invertible transformations such as translations and reparameterizations that preserve the order of the values of variables $\vec{X}$ and of an output $Y$ \citep{Kullback:1959,VergaraEstevez:2014}.

For illustration purposes, only the case with two variables $X$ and an output $Y$ is discussed in the theoretical section. However, a generalization to multiple variables can be derived directly from the independence assumption of random variables, as will be done in later sections.

\subsection{Mutual information}
\label{sec:mutual_information}

Mutual information \citep{ShannonWeaver:1949,CoverThomas:2006} relates the joint probability distribution $p(x, y)$ of two discrete random variables with the product of their marginal distribution $p(x)$ and $p(y)$,
\begin{align}
    I(Y; X) & = \sum_{y \in Y} \sum_{x \in X} p(y, x) \log \frac{p(y, x)}{p(y) p(x)} \notag \\
    & \equiv D_\text{KL}(p(y, x) \| p(y) p(x)) \ .
    \label{eq:mutual_information}
\end{align}
Mutual information is non-negative, is zero if and only if the variables are statistically independent, $p(x, y) = p(x) p(y)$ (independence assumption of random variables), and increases monotonically with the mutual interdependence of variables otherwise. Further, mutual information indicates the reduction in the uncertainty of $Y$ given $X$ as $I(Y; X) = H(Y) - H(Y \vert X)$, where $H(Y)$ denotes the Shannon entropy and $H(Y \vert X)$ the conditional entropy \citep{Shannon:1948}. Shannon entropy $H(Y)$ is defined as the expected value of the negative logarithm of the probability density $p(y)$,
\begin{equation}
    H(Y) = - \sum_{y \in Y} p(y) \log p(y) \ ,
    \label{eq:shannon_entropy}
\end{equation}
and can be interpreted as a measure of the uncertainty on the occurrence of events $y$ whose probability density $p(y)$ is described by the random variable $Y$.

The conditional entropy $H(Y \vert X)$ quantifies the amount of uncertainty about the value of $Y$, provided the value of $X$ is known. It is given by
\begin{equation}
    H(Y \vert X) = - \sum_{y \in Y} \sum_{x \in X} p(y, x) \log p(y \vert x) \ ,
\end{equation}
where $p(y \vert x) = p(y, x) / p(x)$ is the conditional probability of $y$ given $x$. Clearly, $0 \leq  H(Y \vert X) \leq H(Y)$ with $H(Y \vert X) = 0$ if variables $X$ and $Y$ are related by functional dependency and $H(Y \vert X) = H(Y)$ if variables are independent of each other.

Although mutual information is restricted to the closed interval $0 \leq I(Y; X) \leq H(Y)$, the upper bound is still dependent on $Y$. To facilitate comparisons, mutual information is normalized,
\begin{equation}
    D(Y; X) = \frac{I(Y; X)}{H(Y)} = \frac{H(Y) - H(Y \vert X)}{H(Y)} \ .
    \label{eq:fraction_of_information}
\end{equation}
Normalized mutual information, hereafter referred to as fraction of information (also known as coefficients of constraint \citep{CoombsDawesEtAl:1970}, uncertainty coefficient \citep{PressFlanneryEtAl:1989}, or proficiency \citep{WhiteSteingoldEtAl:2004}) quantifies the proportional reduction in uncertainty about $Y$ when $X$ is given. It is in the range $D: \mathbb{R} \times \mathbb{R} \to [0, 1]$, where 0 and 1 represent statistical independence and functional dependence, respectively \citep{ReimherrNicolae:2013}.

\subsection{Probability and cumulative distributions}
\label{sec:limitations_of_mutual_information}

Mutual information and fraction of information are only defined for discrete distributions. Although mutual information can be generalized to continuous distributions,
\begin{equation}
    I(Y; X) = \int_{y \in Y} \int_{x \in X} p(y, x) \log \frac{p(y, x)}{p(y) p(x)}\,dx\,dy \ ,
\end{equation}
probability densities are not always accessible from sample data and therefore need to be estimated. Common algorithms for probability-density estimations are clustering \citep{NguyenMuellerEtAl:2013,PfitznerLeibbrandtEtAl:2008,XuTian:2015}, discretization \citep{FayyadIrani:1993,DoughertyKohaviEtAl:1995,NguyenMuellerEtAl:2014}, and density estimation \citep{KellerMullerEtAl:2012,Garcia:2010,BernacchiaPigolotti:2011,OBrienCollinsEtAl:2014,OBrienKashinathEtAl:2016}. However, these methods implicitly introduce adjustable parameters whose choice has a strong impact on the strength of mutual dependence between a set of variables $\vec{X} = \{ X_1, \ldots, X_n \}$ and an output $Y$, and thus on the ranking induced by the relevance function and a feature-selection criterion (cf., Fig.~\ref{fig:cumulative_entropy_visualization}). In practice, such approaches are extremely dependent on the applied parameter set and therefore are sensitive to the scale of variables (cf., Section~\ref{sec:experiments}).

An alternate approach to using probability distributions is cumulative probability distributions to determine the mutual dependence between variables. Cumulative probability distributions $P$ (and residual cumulative distribution $P' \approx 1 - P$) of a variable $X$ evaluated at $x$ describe the probability that $X$ takes on a value less than or equal to $x$ (or a value greater than or equal to $x$, respectively),
%
\begin{align}
    P(x) & := P(X \leq x) \ , \\
    P'(x) & := P(X \geq x) = 1 - P(X < x) \ .
    \label{eq:cumulative-and-resiudal-distributions}
\end{align}
If the derivatives exists, they are the anti-derivatives of probability distributions,
\begin{equation}
    P(x) := P(X \leq x) = \int_{-\infty}^x p(x')\,dx' \ .
    \label{eq:cumulative_probability_distribution}
\end{equation}
Both residual and cumulative distributions are defined for continuous and discrete variables and are based on accumulated statistics. As such, they are more regular and less sensitive to statistical noise than probability distributions \citep{CrescenzoLongobardi:2009a,CrescenzoLongobardi:2009b}. In particular, they are monotonically increasing and decreasing, respectively, i.e., $P(x_1) \leq P(x_2)$ or $P'(x_1) \geq P'(x_2)$, $\forall x_1 \leq x_2$, with limits
\begin{equation}
    \begin{array}{r}
        \displaystyle\lim_{x \to -\infty} P(x) = 0\\
        \displaystyle\lim_{x \to \infty} P(x) = 1
    \end{array}\ , \quad \begin{array}{r}
        \displaystyle\lim_{x \to -\infty} P'(x) = 1\\
        \displaystyle\lim_{x \to \infty} P'(x) = 0
    \end{array} \ .
\end{equation}
Similar to probability distributions, cumulative and residual cumulative distributions are invariant under a change of variables. However, they are invariant only for parameterizations that preserve the order of the values of each variable $X \in \vec{X}$ and $Y$. Positive monotonic transformations $\mathcal T: \mathbb{R} \to \mathbb{R}$,
\begin{multline}
    P(x) = P(\mathcal T(x)) \quad \forall x \in X: x \mapsto \mathcal T(x) \\
    \quad \text{such that} \quad x_1 < x_2 \Rightarrow \mathcal T(x_1) < \mathcal T(x_2) \ ,
    \label{eq:positive_monotonic_transformations}
\end{multline}
such as translations and nonlinear scaling of variables are among the transformations whose cumulative distributions remain invariant. In contrast, invertible and especially non-invertible mappings \citep{Mira:2007} (such as inversions, $\mathcal T(X) \mapsto \pm X$, and non-bijective transformations, e.g., $\mathcal T(X) = \pm \lvert X \rvert$) change the order of the values of a variable and with it the cumulative distribution. Consequently, these mappings must be considered as additional variables during feature selection if it is expected that such transformations might be related to an output.

\subsection{Cumulative mutual information}
\label{sec:cumulative_mutual_information}

Cumulative mutual information is an alternative measure of uncertainty that extends Shannon entropy (and mutual information) to random variables of continuous distributions. Cumulative mutual information has the same properties as mutual information to be monotonically increasing with the cardinality of the set of variables (Eq.~\ref{eq:monotonicity_information_theory}). Analogous to mutual information, cumulative mutual information describes the inherent dependence expressed in the joint cumulative distribution $P(x, y) = P(X \leq x, Y \leq y)$ of random variables $x \in X$ and $y \in Y$ relative to the product of their marginal cumulative distribution $P(x)$ and $P(y)$,
\begin{align}
    \mathcal I(Y; X) & = \int_{y \in Y} \int_{x \in X} P(y, x) \log \frac{P(y, x)}{P(y) P(x)} \,dx\,dy \notag \\
    & = D_\text{KL}(P(y, x) \| P(y) P(x)) \ .
    \label{eq:cumulative_mutual_information}
\end{align}
The independence assumption of random variables, $P(y, x) = P(y) P(x)$, induces a measure that is again zero only if variables $X$ and $Y$ are statistically independent, and non-negative otherwise. Similarly to mutual information, cumulative mutual information quantifies the degree of dependency as the reduction in the uncertainty of $Y$ given $X$, i.e., $\mathcal I(Y; X) = \mathcal H(Y) - \mathcal H(Y \vert X)$. It is a function of cumulative entropy $\mathcal H(Y)$ and conditional cumulative entropy $\mathcal H(Y \vert X)$,
\begin{align}
    \mathcal H(Y) & = - \int_{y \in Y} \int_{x \in X} P(y, x) \log P(y) \,dx\,dy \\
    \mathcal H(Y \vert X) & = - \int_{y \in Y} \int_{x \in X} P(y, x) \log P(y \vert x) \,dx\,dy \ ,
    \label{eq:cumulative_and_conditional_cumulative_entropy}
\end{align}
where $P(y \vert x) = P(y, x) / P(x)$ is the conditional cumulative distribution of $Y \leq y$ given $X \leq x$ (cf., Tab.~\ref{tab:notation}). Again, $\mathcal H(Y \vert X) = 0$ if variables $X$ and $Y$ are functional dependent and $\mathcal H(Y \vert X) = \mathcal H(Y)$ if variables $X$ and $Y$ are independent of each other.

Bounds restrict cumulative mutual information to a closed interval $0 \leq \mathcal I(Y; X) \leq \mathcal H(Y)$ with an upper bound dependent on $Y$. For this reason, cumulative mutual information is normalized,
\begin{equation}
    \mathcal D(Y; X) = \frac{\mathcal I(Y; X)}{\mathcal H(Y)} = \frac{\mathcal H(Y) - \mathcal H(Y \vert X)}{\mathcal H(Y)} \ ,
    \label{eq:fraction_of_cumulative_information}
\end{equation}
and, likewise to mutual information, is hereafter referred to as fraction of cumulative mutual information.

\section{Empirical estimations of cumulative entropy and cumulative mutual information}
\label{sec:empirical_estimations_of_cumulative_entropy_and_cumulative_mutual_information}

The closed-form expression of cumulative mutual information (Eq.~\ref{eq:cumulative_mutual_information}) quantifies the dependence of a set of variables based on the assumption of smooth and differentiable cumulative distributions. Due to the limited availability of data, however, the exact functional shape of the cumulative distribution is not directly accessible and hence must be empirically inferred from a limited set of sample data.

For this reason, let us assume an empirical sample $\{(y_1, x_1)$, $(y_2, x_2)$, $\ldots$, $(y_n, x_n)\}$ drawn independently and identically distributed (i.i.d.)\ according to the joint distribution of $X$ and $Y$. Such sample data induces empirical (cumulative) probability distributions for all variables $Z \in \{Y, X\}$, which lead to empirical estimates $\hat{\mathcal E}$ of an estimator ${\mathcal E}$ (cf., Tab.~\ref{tab:notation}).

Based on the maximum likelihood estimate \citep{Dutta:1966,Rossi:2018}, the cumulative probability distribution $\hat{P}(Z \leq z)$ can be obtained by counting the frequency of occurring values of a variable $Z$:
\begin{multline}
    \hat{P}(Z \leq z) = \frac{1}{n} \sum_{i = 1}^n \mathbf{1}_{z_i \leq z} = \frac{1}{n} \left \lvert \{\,i \mid z_i \leq z\,\} \right \rvert \ ,\\
    \forall z_i \in Z \ , \quad z \in Z,\ Z \in \{ Y, X \} \ ,
    \label{eq:maximum-likelihood-estimator}
\end{multline}
where $\mathbf{1}_A$ denotes the indicator function that is one if $A$ is true, and zero otherwise. Equation~\ref{eq:maximum-likelihood-estimator} asymptotically converges to $P(Z \leq z)$ as $n \to \infty$ for every value of $z \in Z$ (Glivenko-Cantelli theorem: \cite{Glivenko:1933,Cantelli:1933}). Thus, any empirical estimate, $\hat{\mathcal E}$, based on empirical cumulative distributions converges pointwise as $n \to \infty$ to the actual value of $\mathcal E$, i.e., $\hat{\mathcal E}(Z) \to \mathcal E(Z)$ \citep{RaoChenEtAl:2004,CrescenzoLongobardi:2009b}.

\subsection{Empirical cumulative entropy}
\label{sec:empirical_cumulative_entropy}

For i.i.d.\ random samples that contain repeated values, the maximum likelihood estimate of the cumulative entropy $\mathcal H$ (Eq.~\ref{eq:cumulative_and_conditional_cumulative_entropy}) can be obtained by calculating the empirical cumulative distribution $\hat{P}$ according to Eq.~\ref{eq:maximum-likelihood-estimator},
\begin{equation}
    \begin{aligned}
        \hat{\mathcal H}(Y) & = - \sum_{i = 1}^{k - 1} \Delta y_i \hat{P}(y) \log \hat{P}(y) \\
        & = - \sum_{i = 1}^{k - 1} \left ( y_{(i + 1)} - y_{(i)} \right ) \frac{n_i}{n} \log \frac{n_i}{n} \ ,
    \end{aligned}
    \label{eq:emiprical_cumulative_entropy}
\end{equation}
where $y_{(i)}$ denotes the values $y_{(0)} < y_{(1)} < \cdots < y_{(k)}$ occurring in the data set in sorted order of $Y$ with $y_{(0)} = -\infty$, multiplicity $n_i = \left \lvert \{ j \in n: y_{(i - 1)} < y_j \leq y_{(i)}\} \right \rvert$, and constraint $n = \sum_{i = 1}^k n_i$.

\subsection{Empirical conditional cumulative entropy}
\label{sec:empirical_conditional_cumulative_entropy}

Similar to empirical cumulative entropy, conditional cumulative entropy can be estimated by
\begin{align}
    \hat{\mathcal H}(Y; \vec{X}) & = - \sum_{i = 1}^{n - 1} \sum_{j = 1}^{n - 1} \Delta y_i \Delta \vec{x}_j \hat{P}(y_i, \vec{x}_j) \log \hat{P}(y_i \vert \vec{x}_j) \notag \\
    & = - \sum_{i = 1}^{n - 1} \sum_{j1 = 1}^{n - 1} \ldots \sum_{jd = 1}^{n - 1} \bigl ( y_{i + 1} - y_i \bigr ) \bigl ( x^{(1)}_{j1 + 1} - x^{(1)}_{j1} \bigr ) \cdot \notag \\
    & \qquad \ldots \cdot \bigl ( x^{(d)}_{jd + 1} - x^{(d)}_{jd} \bigr ) \hat{P}(y_i, \vec{x}_j) \log \hat{P}(y_i \vert \vec{x}_j) \ ,
    \label{eq:empirical_conditional_cumulative_entropy}
\end{align}
where $\hat{P}(y_i, \vec{x}_j)$ denotes the joint cumulative distribution of $y_i \in Y$, $\vec{x}_j \in \vec{X}$, $\vec{X} = \{ X_1, \ldots X_d \}$, and $x_i^{(k)} \in X_k$ is the $i$ component of the $k$-th variable of the data set ($k = 1, \ldots d$). In contrast to the empirical cumulative entropy, which can be calculated from the set of sample data with linear time complexity $\mathcal O(n)$, the empirical conditional cumulative entropy has exponential time complexity $\mathcal O(n^d)$. The non-parametric estimation of the joint or conditional cumulative distribution therefore becomes computationally demanding for data sets with a large number $d$ of variables and data samples $n$.

\subsection{Empirical cumulative mutual information}
\label{sec:empirical_cumulative_mutual_information}

By construction, cumulative entropy is sensitive to the range of $Y$ (cf., Eq.~\ref{eq:emiprical_cumulative_entropy}). The same is true for conditional cumulative entropy $\mathcal H(Y \vert X) $ and its empirical estimate $\hat{\mathcal H}(Y \vert X)$ (cf., Eq.~\ref{eq:empirical_conditional_cumulative_entropy}). Fraction of cumulative mutual information ${\mathcal D}$, e.g., the ratio between cumulative entropy and conditional cumulative entropy, is independent of the scale of $X$ and $Y$ (Eq.~\ref{eq:fraction_of_cumulative_information}). Formally, its empirical estimate $\hat{\mathcal D}$ is given by
\begin{multline}
    \hat{\mathcal D}(Y; \vec{X}) = 1 - \frac{1}{n} \Biggl [ \sum_{i, j = 1}^{n - 1} \Delta y_i \Delta x_j \hat{P}(y_i, \vec{x}_j) \log \hat{P}(y_i \vert \vec{x}_j) \Bigm /\ \\
    \sum_{i,j = 1}^{n - 1} \Delta y_i  \Delta x_j \hat{P}(y_i) \log \hat{P}(y_i) \Biggr ] \ .
    \label{eq:empirical_fraction_of_cumulative_information}
\end{multline}
Computationally, we apply the following trick: To eliminate the implicit scale dependence of $X$, we use the fact that variables $X$ are invariant under rank-order preserving transformations $\mathcal T$ (Eq.~\ref{eq:positive_monotonic_transformations}). Then, all variables can be scaled to $x' = \mathcal T(x)$ such that $\Delta x_i' = x_{i + 1} - x_i$ is constant and the volume element $dx'$ in the integrals cancels out (cf., Eq.~\ref{eq:cumulative_and_conditional_cumulative_entropy}).
\begin{equation}
    \hat{\mathcal D}(Y; X) = 1 - \frac{1}{n} \sum_{j = 1}^{n - 1} \frac{\sum_{i = 1}^{n - 1} \Delta y_i P(y_i, x_j) \log P(y_i \vert x_j)}{\sum_{i = 1}^{n - 1} \Delta y_i P(y_i, x_j) \log P(y_i)} \ .
    \label{eq:empirical_fraction_of_cumulative_mutual_information}
\end{equation}
Such a transformation is always possible and effectively removes the implicit range dependence of variables from the fraction of cumulative information in the computation.

\section{Baseline adjustment}
\label{sec:baseline_adjustment}

The limited availability of data makes it challenging to estimate or calculate dependencies on empirical estimators. Because measures are meant to provide a comparison mechanism, empirical estimators need to assign a value (dependence score) close to zero for statistical independent variables and a score close to one for functional dependent variables. However, empirical estimators based on mutual information are known to never reach their theoretical maximum (functional dependence) or minimum (statistical independence), respectively, and are known to assign stronger dependences for larger sets of variables regardless of the underlying relationship \citep{FoucheBoehm:2019,FoucheMazankiewiczEtAl:2021,VinhEppsEtAl:2009,VinhEppsEtAl:2010}. Consequently, measures based on mutual information have a considerable inherent bias and therefore may incorrectly identify variables as relevant that are not related to an output $Y$. To actually compare dependence measures between subsets and different sizes of variable sets, an adjustment to mutual information is necessary. One solution to estimate the relevance of a set of variables $X$ and $Y$ is to compare the relevance of a variable $X$ and an output $Y$ to the mean $\hat{\mathcal E}_0$ of an empirical estimator $\hat{\mathcal E}$,
\begin{equation}
    \hat{\mathcal E}^*(Y; X) = \hat{\mathcal E}(Y; X) - \hat{\mathcal E}_0(Y; X) \ .
    \label{eq:adjusted_measure}
\end{equation}

The mean $\hat{\mathcal E}_0$ requires to be constant across random permutations of all variables independently for each data sample, i.e.,
\begin{equation}
    \hat{\mathcal E}_0(Y; X) := \frac{1}{\lvert \mathcal M \rvert} \sum_{M \in \mathcal M} \hat{\mathcal E}(Y_M; X_M) \ ,
\end{equation}
where $M \in \mathcal M$ is a specific realization of such a permutation. The underlying intuition is that the actual value of an empirical estimator $\hat{\mathcal E}$ may be caused by spurious (random) dependences. Therefore, by considering all random permutations of all variables independently for each data sample, the spurious contribution of the empirical estimator can be factored out and an adjusted unbiased empirical estimator obtained. The permutations can be computed by enumeration, which however is impractical. An alternative description is provided by a hypergeometric model of randomness \citep{VinhEppsEtAl:2009,RomanoBaileyEtAl:2014} (also known as permutation model \citep{Lancaster:1969}). Such a model describes the permutation of variables as (cumulative) probability distributions, where the average can be calculated separately for each sample of a data set with quadratic complexity. Under the independence assumption of random variables \citep{VinhEppsEtAl:2009,VinhEppsEtAl:2010}, we derived the correction term for cumulative mutual information as follows,
\begin{multline}
    \hat{\mathcal I}_0(Y; X) = - \sum_{i = 1}^{r - 1} \sum_{j = 1}^{c} \sum_{n_{ij}} \Delta y_i(n_{ij}, a_i, b_j \vert M ) \cdot \\
    \frac{n_{ij}}{n} \log \frac{n_{ij}}{b_j} \mathcal P(n_{ij}, a_i, b_j \vert M) \ ,
    \label{eq:expected_cumulative_mutual_information}
\end{multline}
where the difference $\Delta y_i(M)$ between two consecutive values of $Y$ can be described by a binomial distribution,
\begin{equation}
    \Delta y_i(n_{ij}, a_i, b_j \vert  M) = \frac{1}{\mathcal N} \sum_{k = 1}^{k_\text{max}} \binom{r - k - 1}{b_j - 2} \bigl ( y_{(i + k)} - y_{(i)} \bigr ) \ ,
\end{equation}
$k_\text{max}$ is the upper limit is given by $k_\text{max} = \min(n - b_j + 1, r - i)$. $\mathcal N$ is a normalization constant,
\begin{equation}
    \mathcal N = \sum_{k = 1}^{k_\text{max}} \binom{r - k - 1}{b_j - 2} \ ,
\end{equation}
and $\mathcal P(n_{ij}, a_i, b_j \vert  M)$ is the probability to encounter an associative cumulative contingency table subject to fixed marginals between all permutations of two variables $X$ and $Y$ with $\lvert Y_i \rvert = a_i$, $i = 1, \ldots, r$ and $\lvert X_j \rvert = b_j$, $j = 1, \ldots, c$. $n_{ij}$ is a specific realization of the joint cumulative distribution $P(y_i, x_j)$ given row marginal $a_i$ and column marginal $b_j$. The details can be found in the appendix and are analogous to the baseline adjustment for mutual information \citep{VinhEppsEtAl:2009}.

The empirical estimator $\hat{\mathcal E}_0$ in Eq.~\ref{eq:adjusted_measure} is required to vanish for a large number of samples $\hat{\mathcal E}_0(Y; X)~\to~0$ as $n~\to~\infty$ in case there is an exact functional dependence between $X$ and $Y$ \citep{RomanoVinhEtAl:2016}. Further, $\hat{\mathcal E}_0$ is required to be zero if variables are proportional to the output, $\mathcal E_0(Y; X)~\to~0$ as $X~\to~Y$.

In practice, $\hat{\mathcal E}_0$ is generally greater than zero when the number of data samples is limited and can become as large as $\hat{\mathcal E}$ when the number of data samples is very small. $\hat{\mathcal E}_0$ can therefore be interpreted as a correction term for comparing empirical estimates of different sets of variables on a common baseline: In general, if the value of the correction term is large, more data samples are needed to reliably estimate the dependence between $X$ and $Y$. If the value of correction term is small, the adjusted empirical estimator either indicates a strong mutual dependence between $X$ and $Y$ (high $\hat{\mathcal E}^*$) or a weak mutual dependence, if the variables of the data set are not related to $Y$ (low $\hat{\mathcal E}^*$).

For cumulative mutual information, we define the empirical estimator as follows
\begin{align}
    \hat{\mathcal I}^*(Y; X) & = \hat{\mathcal I}(Y; X) - \hat{\mathcal I}_0(Y; X) \ , \label{eq:adjusted_fraction_of_cumulative_mutual_information} \\
    \hat{\mathcal D}^*(Y; X) & = \frac{\hat{\mathcal I}^*(Y; X)}{\hat{\mathcal H}^*(Y)} = \hat{\mathcal D}(Y; \vec{X}) - \hat{\mathcal D}_0(Y; \vec{X}) \ , \label{eq:adjusted_fraction_empirical_cumulative_information}
\end{align}
where $\hat{\mathcal I}^*(Y; X)$ is the adjusted empirical cumulative mutual information, $\hat{\mathcal D}^*(Y; X)$ is the adjusted fraction of empirical cumulative information, and $\hat{\mathcal I}_0(Y; X)$ is the expected cumulative mutual information under the independence assumption of random variables.

\section{Total cumulative mutual information}
\label{sec:total_cumulative_mutual_information}

Empirical cumulative mutual information provides a non-parametric deterministic measure to estimate the dependence of continuous distributions. Equation~\ref{eq:fraction_of_cumulative_information} estimates cumulative mutual information based on cumulative probability distributions, $P(X) = P(X \geq x)$. Similarly, a measure can be instantiated for residual cumulative probability distributions, $P'(X) := P(X \geq x) = 1 - P(X \leq x)$,
\begin{equation}
    \mathcal D'(Y; X) = \frac{\mathcal H'(Y) - \mathcal H'(Y \vert X)}{\mathcal H'(Y)} \ .
    \label{eq:empirical_fraction_of_residual_cumulative_mutual_information}
\end{equation}
Both measures $\mathcal D(Y; X)$ and $\mathcal D'(Y; X)$ estimate the dependence between a set of variables and an output from different sides of the distribution: therefore, they set lower and upper bounds on the information they contain. As the sample size increases to infinity, both measures converge to the same value. However, due to the limited number of data samples (cf., Section~\ref{sec:baseline_adjustment}), these measures are different and need to be adjusted in practice,
\begin{align}
    \hat{\mathcal D}^*(Y; X) & = \hat{\mathcal D}(Y; X) - \hat{\mathcal D}_0(Y; X) \ , \notag \\
    \hat{\mathcal D}^{*\prime}(Y; X) & = \hat{\mathcal D}'(Y; X) - \hat{\mathcal D}_0'(Y; X) \ .
    \label{eq:adjusted_empirical_fraction_of_cumulative_mutual_information}
\end{align}
The baseline adjustment turns both measures convex by relating the strength of a dependence among variables with the dependence of the same set of variables under the independence assumption of random variables \citep{VinhEppsEtAl:2009,VinhEppsEtAl:2010}. They can therefore be used to efficiently search for the strongest mutual dependence between a set of variables and an output, e.g., by using the minimum contribution of fraction of empirical cumulative mutual information of the two measures,
\begin{equation}
    \hat{\mathcal D}_\text{min}^*(Y; X) := \min ( \hat{\mathcal D}^*(Y; X), \hat{\mathcal D}^{*\prime}(Y; X)) \ .
    \label{eq:total_cumulative_mutual_information}
\end{equation}
Total cumulative mutual information (TCMI) combines $\hat{\mathcal D}^*(Y; X)$ and $\hat{\mathcal D}^{\prime*}(Y; X)$ into a single measure. TCMI is defined as the average strength of cumulative mutual dependence between a set of variables $X$ and an output $Y$,
\begin{equation}
    \langle \hat{\mathcal D}_\text{TCMI}^*(Y; X) \rangle := \langle \hat{\mathcal D}_\text{TCMI}(Y; X) \rangle - \langle \hat{\mathcal D}_\text{TCMI, 0}(Y; X) \rangle \ ,
    \label{eq:average_total_cumulative_mutual_information}
\end{equation}
where
\begin{align}
    & \langle \hat{\mathcal D}_\text{TCMI}(Y; X) \rangle = \frac{1}{2} \left [ \hat{\mathcal D}(Y; X) + \hat{\mathcal D'}(Y; X) \right ] \notag \\
    & \langle \hat{\mathcal D}_\text{TCMI, 0}(Y; X) \rangle = \frac{1}{2} \left [ \hat{\mathcal D}_0(Y; X) + \hat{\mathcal D}_0'(Y; X) \right ] \ .
\end{align}

\section{Feature selection}
\label{sec:feature_selection}

Feature selection (Eq.~\ref{eq:feature_selection}) is an optimization problem that either requires a convex dependence measure or additional criteria to judge the optimality of a feature set \citep{YuPrincipe:2019}. Measures based on (cumulative) mutual information do not meet either requirement, but an adjusted measure such as TCMI does.

As already mentioned in the introduction, the optimal search strategy (subset selection) of $k$ features from an initial set of variables $\vec{X} = \{ X_1, \ldots, X_d \}$ is a combinatorial and exhaustive search procedure that is only applicable to low-dimensional problems. An efficient alternative to the exhaustive search is the (depth-first) branch-and-bound algorithm \citep{LandDoig:1960,NarendraFukunaga:1977,Clausen:1999,MorrisonJacobsonEtAl:2016}. The branch-and-bound algorithm guarantees to find an optimal set of feature variables without evaluating all possible subsets. The performance depends crucially on the variables of a data set and the maximum strength of the mutual dependence between a set of variables and an output. It may be that an output is only weakly related to the variables in the data set, making it necessary to repeat the feature selection with a different set of variables. It may also be that prior knowledge of potential mutual dependences is available, which speeds up the feature selection (e.g., that only $m < k$ of $d$ variables $X$ are related to $Y$ and therefore not all combinations need to be implicitly enumerated).

The branch-and-bound algorithm maximizes an objective function $\mathcal Q^*: \vec{X}' \to \mathbb{R}$ defined on a subset of variables $\vec{X}' \subseteq \vec{X}$ by making use of the monotonicity condition of a feature-selection criterion, $\mathcal Q: \vec{X}' \to \mathbb{R}$, and a bounding criterion, $\bar{\mathcal Q}: \vec{X}' \to \mathbb{R}$. The monotonicity condition requires that feature subsets $\vec{X}_1$, $\vec{X}_2$, $\cdots$, $\vec{X}_k$, $k=1,\ldots,d$, obtained by sequentially adding $k$ features from the set of variables $\vec{X}$, satisfy
\begin{equation}
    \vec{X}_1 \subseteq \vec{X}_2 \subseteq \cdots \subseteq \vec{X}_k \ , \qquad \vec{X}_k \subseteq \vec{X} \ ,
\end{equation}
so that the feature-selection criterion $\mathcal Q$ and bounding criterion $\bar{\mathcal Q}$ are monotonically increasing and decreasing respectively,
\begin{equation}
    \begin{aligned}
        & \mathcal Q(\vec{X}_1) \leq \mathcal Q(\vec{X}_2) \leq \cdots \leq \mathcal Q(\vec{X}_k) \\
        & \bar{\mathcal Q}(\vec{X}_1) \geq \bar{\mathcal Q}(\vec{X}_2) \geq \cdots \geq \bar{\mathcal Q}(\vec{X}_k) \ .
    \end{aligned}
    \label{eq:branch_and_bound_monotonicity_condition}
\end{equation}

\begin{figure*}
    \centering
    \includegraphics[width=\textwidth]{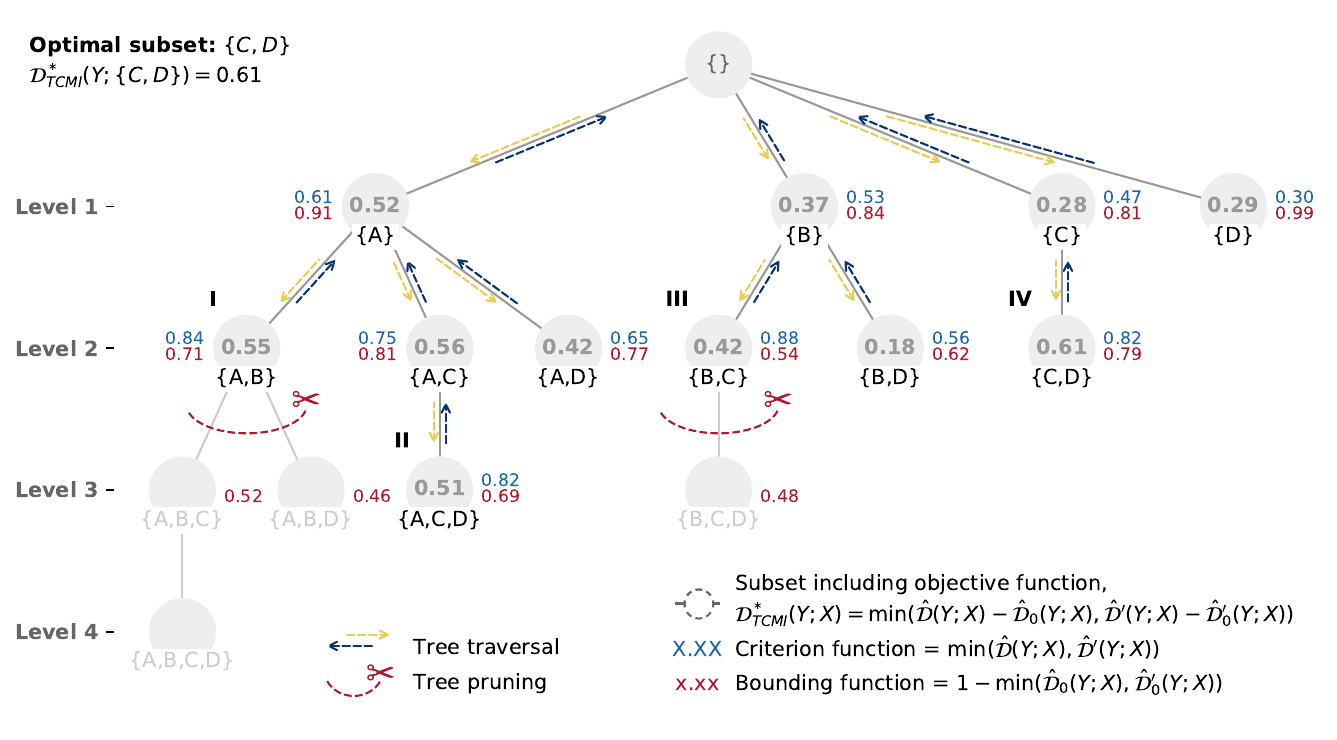}
    \caption{Example of a depth-first tree search strategy of the branch-and-bound algorithm \citep{LandDoig:1960,NarendraFukunaga:1977,Clausen:1999,MorrisonJacobsonEtAl:2016} to search for the optimal subset of features. Shown is the tree traversal going from top to down and left to right by dashed arrows, the estimated fraction of total cumulative information (objective function inside circles), subsets of features (labels at the bottom of the circles), fraction of cumulative information (criterion function, first number, right or left the circles), and the expected fraction of cumulative information contribution (bounding function, second number, right or left the circles). Capital roman symbols indicate applied pruning rules or updates of the current maximum objective function. Anytime the objective function in some internal nodes exceeds the bounding function of sub-trees (I), it decreases (II) -- either due to the condition Eq.~\ref{eq:branch_and_bound_monotonicity_condition} or the bounding function is lower than the current maximum value of the objective function (III), sub-trees can be pruned and computations be skipped. On termination of the algorithm, the bound contains the optimum objective function value (IV).}
    \label{fig:branch_and_bound_algorithm}
\end{figure*}

The branch-and-bound algorithm builds a search tree of feature subsets $\vec{X}' \subseteq \vec{X}$ with increasing cardinality \citep{Clausen:1999,MorrisonJacobsonEtAl:2016} (Alg.~\ref{alg:branch_and_bound_algorithm} and Fig.~\ref{fig:branch_and_bound_algorithm}). Initially the tree contains only the empty subset (the root node). At each iteration, a limited number of (non-redundant) sub-trees are generated by augmenting one variable $X \in \vec{X}$ at a time to the current subset and then adding it to the search tree (branching step). While traversing the tree from the root down to terminal nodes from left to right, the algorithm keeps the information about the currently best subset $X^* := \vec{X}_k$ and the corresponding objective function it yields (the current maximum). Anytime the objective function $\mathcal Q^*$ in some internal nodes exceeds the bounding criterion $\bar{\mathcal Q}$ of sub-trees, it decreases (either due to the condition Eq.~\ref{eq:branch_and_bound_monotonicity_condition} or the bounding criterion is lower than the current maximum value of the objective function), sub-trees can be pruned and computations be skipped (bounding step). Once the entire tree has been examined, the search terminates and the optimal set of variables is returned, along with a ranking of sub-optimal variable sets in descending order of the value of the objective function values.
\begin{algorithm}[t]
    \begin{algorithmic}[1]
        \Function{branch\_and\_bound}{features $\vec{X}$, target $Y$}
            \State $\vec{S}_0$ := $\varnothing$;
            \State Subsets := $\{ \vec{S}_0 \}$;
            \State Optimal := $\vec{S}_0$;

            \While{Subsets}
                \For{$X_i \in \vec{X} \setminus \vec{S}_{k - 1}$}
                    $\vec{S}_k$ := $\vec{S}_{k - 1} \otimes X_i$;
                    \State Compute $\mathcal Q(Y; \vec{S}_k)$ and $\bar{\mathcal Q}(Y; \vec{S}_k)$;
                    \If{$\mathcal Q(Y; \vec{S}_k) < \bar{\mathcal Q}(Y; \vec{S}_k)$}
                        \State Subsets := Subsets $\cup \{ \vec{S}_k \}$;
                        \If{$\mathcal Q(Y; \vec{S}_k) > \mathcal Q(Y; \text{Optimal})$}
                            \State Optimal := $\vec{S}_k$;
                        \EndIf
                    \EndIf
                \EndFor
            \EndWhile
        \State \textbf{return} Optimal
        \EndFunction
    \end{algorithmic}
    \caption{A pseudo-code listing of the branch-and-bound algorithm \citep{Clausen:1999,MorrisonJacobsonEtAl:2016}.}
    \label{alg:branch_and_bound_algorithm}
\end{algorithm}
As objective and criterion function we set
\begin{equation}
    \mathcal Q^* = \mathcal D_\text{TCMI}^*(Y; X) \ ,
\end{equation}
the criterion function to be
\begin{equation}
    \mathcal Q = \min (\mathcal D(Y; X), \mathcal D'(Y; X)) \ ,
\end{equation}
and, as a pruning rule, the bounding criterion to be (cf., Eq.~\ref{eq:total_cumulative_mutual_information}),
\begin{equation}
    \bar{\mathcal Q} = 1 - \min (\hat{\mathcal D}_0(Y; X), \hat{\mathcal D}_0'(Y; X)) \ .
\end{equation}
Proofs for the monotonicity conditions for $\mathcal Q$ and $\bar{Q}$ follow similar arguments as for Shannon entropy \citep{MandrosBoleyEtAl:2017} and are provided in the appendix.

\subsection{Complexity Analysis}

The computational complexity of the branch-and-bound algorithm is largely determined by two factors: the branching factor $B$ and the depth $D$ of the tree \citep{MorrisonJacobsonEtAl:2016}. The branching factor is the maximum number of generated variables combinations at each level $l$ of the tree and can be estimated by the central binomial coefficient $B \leq \max_{l = 1, \ldots, D} \binom{d}{l} \approx \binom{d}{d/2}$, if $\vec{X}$ has $d$ variables. The depth $D$ of the tree is given by the largest cardinality of a variable set, represented as the longest path in the tree from the root to a terminal node. The ranking of the variable sets involves $\mathcal O((n \log n)^d)$ sorting operations when all variables are relevant. Thus, any branch-and-bound implementation has worst-case $\mathcal O(M \cdot B^D)$ computational time complexity, where $M$ is the time needed to evaluate the feature-selection criterion for a combination of variables in the tree.

In the worst case, for $n$ number of example data and $d$ variables, cumulative mutual information requires to evaluate the integral $\mathcal O(n^d)$ times and $\mathcal O(n^2)$ times to calculate the baseline adjustment term. Thus, TCMI has time complexity $M \sim \mathcal O(n^d)$ and a feature-subset search in the current implementation suffers from the curse of dimensionality \citep{KollerSahami:1996}.

As a result, the total time complexity of the feature selection algorithm is non-deterministic polynomial-time (NP)-hard and, in general, the search strategy of examining all possible subsets is not viable. In the vast majority of cases, however, dependencies are relatively simple relationships of only a small number of features. In addition, feature selection can be restricted at any time to examine subsets of variables that are less than or equal to a predefined dimensionality. Then the time complexity is greatly reduced and the feature selection can be solved in polynomial time. Whether the assumptions apply to arbitrary data sets is a case-by-case study. However, indicators such as the convergence rate of the TCMI approaching the maximum value or the estimated strength of the relationships are helpful in exploratory data analysis to search for the relevant features of a data set.

\section{Experiments}
\label{sec:experiments}

To demonstrate the performance of TCMI in different settings, we first consider generated data and show that our method can detect both univariate and multivariate dependences. Then, we discuss applications of TCMI on data sets from the KEEL and UCI Machine Learning Repository \citep{Alcala-FdezSanchezEtAl:2009,Alcala-FdezFernandezEtAl:2011,DuaGraff:2017} and a typical scenario from the materials-science community, namely to predict the crystal structure of octet-binary compound semiconductors \citep{GhiringhelliVybiralEtAl:2015,GhiringhelliVybiralEtAl:2017}.

\subsection{Case study on generated data}
\label{sec:generated_data}

In a number of experiments, we test the theoretical properties of TCMI, i.e., its invariance properties and performance statistics. We also study an exemplified feature-selection task to find a bivariate normal distribution embedded in a multi-dimensional space.

\subsubsection{Interpretability of TCMI}
\label{sec:experments_interpretability_tcmi}

\begin{table*}[t]
    \centering
    \begin{tabular}{lc ccc cccc}
        \toprule
        \textbf{Distribution} & $\rho^2$ & $\langle \hat{\mathcal{D}}_\text{TCMI}^*\rangle$ & $\langle \hat{\mathcal{D}}_\text{TCMI} \rangle$ & $\langle \hat{\mathcal{D}}_\text{TCMI, 0} \rangle$ & CMI & MAC & UDS & MCDE \\
        \midrule\normalsize
        \includegraphics[height=1.2em]{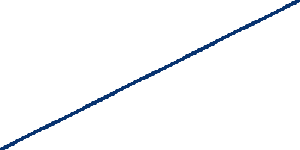} linear & 1.0000 & 0.97 & 1.00 & 0.03 & 1.00 & 1.00 & 0.67 & 1.00 \\
        \includegraphics[height=1.2em]{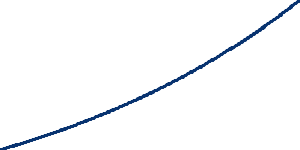} exponential & 1.0000 & 0.97 & 1.00 & 0.03 & 1.00 & 1.00 & 0.65 & 1.00 \\
        \includegraphics[height=1.2em]{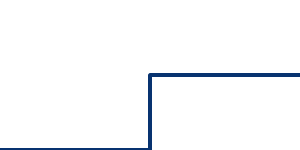} step--2 & 0.9999 & 0.96 & 0.98 & 0.02 & 1.00 & 1.00 & 0.67 & 1.00 \\
        \includegraphics[height=1.2em]{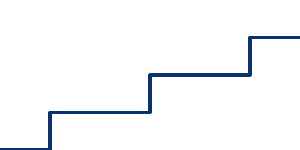} step--4 & 0.9996 & 0.93 & 0.95 & 0.02 & 1.00 & 1.00 & 0.67 & 1.00 \\
        \includegraphics[height=1.2em]{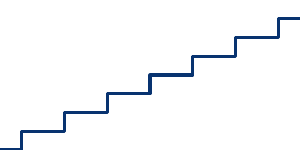} step--8 & 0.9984 & 0.87 & 0.88 & 0.01 & 1.00 & 1.00 & 0.67 & 1.00 \\
        \includegraphics[height=1.2em]{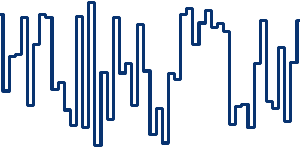} random & 0.0091 & 0.33 & 0.65 & 0.32 & 0.02 & 0.34 & 0.00 & 0.54 \\
        \includegraphics[height=1.2em]{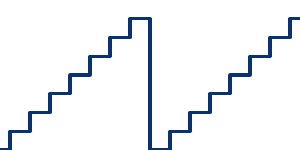} sawtooth--8 & 0.0016 & 0.23 & 0.31 & 0.07 & 0.03 & 0.03 & 0.00 & 0.14 \\
        \includegraphics[height=1.2em]{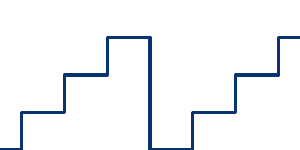} sawtooth--4 & 0.0004 & 0.17 & 0.27 & 0.10 & 0.00 & 0.01 & 0.00 & 0.09 \\
        \includegraphics[height=1.2em]{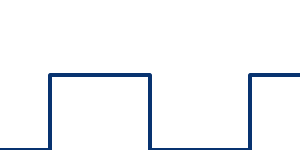} sawtooth--2 & 0.0001 & 0.09 & 0.19 & 0.09 & 0.00 & 0.00 & 0.00 & 0.03 \\
        \includegraphics[height=1.2em]{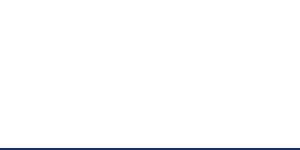} constant & 0.0000 & 0.00 & 0.00 & 0.00 & 0.00 & 0.00 & 0.00 & 1.00 \\
        \bottomrule
    \end{tabular}
    \caption{Dependence scores, $\langle \hat{\mathcal{D}}_\text{TCMI}^*(Y; X) \rangle$, between a linear data distribution and a linear, exponential, step, sawtooth and uniform (random) distribution. The data sample size is $n = 200$. Step-like distributions were generated by discretization of the linear distribution with each value repeating r-times. Sawtooth-like distributions have 2, 4, and 8 number of steps per ramp and $\lceil n / level \rceil$ ramps in total. The table also shows Spearman's rank correlation coefficient squared $\rho^2$, total cumulative mutual information contributions, $\langle \hat{\mathcal{D}}_\text{TCMI}(Y; X) \rangle$ and $\langle \hat{\mathcal{D}}_\text{TCMI, 0}(Y; X) \rangle$, and the scores from similar dependence measures such as CMI \citep{NguyenMuellerEtAl:2013}, MAC \citep{NguyenMuellerEtAl:2014a}, UDS \citep{NguyenMandrosEtAl:2016,WangRomanoEtAl:2017}, and MCDE \citep{FoucheBoehm:2019,FoucheMazankiewiczEtAl:2021}.}
    \label{tab:monotonicity_dependence_measures}
\end{table*}

In the first experiment, we use TCMI, CMI, MAC, UDS, and MCDE to estimate the dependence between a linear data distribution $Y$ of size $n = 200$ and different distributions $X$ as features (Tab.~\ref{tab:monotonicity_dependence_measures}). Besides linear, exponential, and constant distributions (zero vector), we consider stepwise distributions generated by discretizing a linear distribution, where each value is repeated 2, 4, and 8 times. Furthermore, we consider uniform (random) and saw-tooth distributions with 2, 4, and 8 steps per ramp. Results show that (i) the TCMI score increases nonlinearly with the similarity between a variable and an output, (ii) TCMI is zero for a constant distribution, and (iii) TCMI is approaching one for an exact dependence (see also Fig.~\ref{fig:baseline_correction}). CMI, MAC, UDS, and MCDE perform similarly well, but they seem to be less sensitive than TCMI in assessing the strength of a mutual dependence. In particular, the strength of a dependence with CMI, MAC, UDS, MCDE does not change with the shape of a distribution (i.e., of different cumulative probability distributions such as the step-like distributions). MCDE does not differentiate between a linear and a constant distribution, while UDS seems to be limited and does not reach the maximum score even in the presence of an exact dependence.

Due to the limited availability of data samples, a random distribution has a higher TCMI, MAC, and MCDE score, i.e., stronger dependence, than a sawtooth distribution, in agreement with Spearman's rank coefficient of determination $\rho^2$ \citep{Spearman:1904}. It should be noted that the baseline adjustment $\langle \hat{\mathcal{D}}_\text{TCMI, 0} \rangle$ for a random variable is larger than any other tested dependence of Tab.~\ref{tab:monotonicity_dependence_measures}. A large baseline adjustment results in smaller TCMI values, such that it is unlikely that a random variable will be part in any feature selection. However, if the dependences are of the same strength as spurious dependencies induced by random variables, TCMI may select variables that are not related to an output.

\subsubsection{Properties of the baseline correction term}
\label{sec:experiments_properties_baseline_correction}

\begin{figure}[t]
    \centering
    \includegraphics[width=.7\columnwidth]{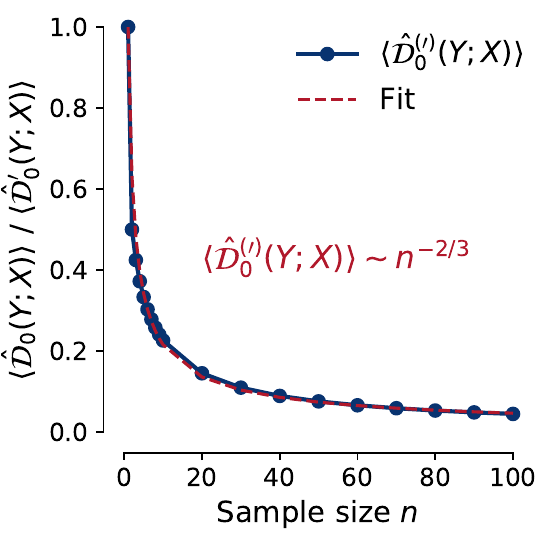}
    \caption{Expected empirical cumulative mutual information, $\langle \hat{\mathcal{D}}_0(Y; X) \rangle$, with respect to the number of sample data. Shown is the dependency (solid line) and a heuristic derived analytic functional relationship (dashed line).}
    \label{fig:baseline_correction}
\end{figure}

In the second experiment (Fig.~\ref{fig:baseline_correction}), we take a closer look at the baseline adjustment term that decreases monotonically with respect to the number of data samples. Baseline adjustment is given by the expected empirical cumulative mutual information (Eqs.~ \ref{eq:expected_cumulative_mutual_information} and \ref{eq:adjusted_fraction_empirical_cumulative_information}). Expected empirical cumulative mutual information follows a clear downward trend in the score with increasing number of sample sizes in all our test cases. For linear dependencies, for example, we found that the baseline adjustment roughly follows a $\langle \hat{\mathcal{D}}^{(\prime)}_0(Y; X) \rangle \sim n^{-2/3}$ scaling law that vanishes as $n \to \infty$ (Fig.~\ref{fig:baseline_correction}). However, the exact scaling behavior in general varies depending on the presence of duplicate values of each variable in a data set.

\subsubsection{Invariance properties of TCMI}
\label{sec:experiment_invariance_properties_tcmi}

In the third experiment, we investigate the invariance properties of TCMI as compared to CMI, MAC, UDS, and MCDE. To this end, we generated random distributions $X$ of different sizes (50, 100, 200, and 500) and reparameterized variables by applying positive monotonic transformations (cf., Section~\ref{sec:limitations_of_mutual_information}). Table~\ref{tab:scale_permutation_invariance} summarizes the results of comparing the dependence scores between a linear distribution and reparameterized variables, e.g., between $\hat{\mathcal D}(Y; X)$ and $\hat{\mathcal D}(Y; \mathcal T(X))$, where monotonic transformations $\mathcal T(X) = a X^k + b$ with $a, b, k \in \mathbb{R}$ and compositions $\mathcal T(X) = \mathcal T_1(X) \pm \cdots \pm \mathcal T_m(X)$ were explored.

\begin{table}[t]
    \centering
    \begin{tabular}{lcc}
        \toprule
        \textbf{Dependence measure} & scale invariant & permutation invariant \\
        \midrule

        CMI  & \xmark & \xmark \\
        MAC  & \cmark & \xmark \\
        UDS  & \xmark & \xmark \\

        MCDE & (\cmark) & (\cmark) \\
        TCMI & \cmark & \cmark \\
        \bottomrule
    \end{tabular}
    \caption{Overview of the invariance properties of the dependence measures: cumulative mutual information (CMI: \cite{NguyenMuellerEtAl:2013}), multivariate maximal correlation analysis (MAC: \cite{NguyenMuellerEtAl:2014a}), universal dependency analysis (UDS: \cite{NguyenMandrosEtAl:2016,WangRomanoEtAl:2017}), Monte Carlo dependency estimation (MCDE: \cite{FoucheBoehm:2019,FoucheMazankiewiczEtAl:2021}), and total cumulative mutual information (TCMI). Checkmarks and crosses in parentheses denote invariance in terms of probabilistic tolerance.}
    \label{tab:scale_permutation_invariance}
\end{table}

By construction, TCMI is invariant under positive monotonic transformations (Eq.~\ref{eq:positive_monotonic_transformations}). Our experiments show that TCMI is indeed both scale and permutation invariant. For CMI, MAC, and UDS, the order of the variables plays a crucial role in determining which permutation of the variable achieves either the highest dependence score (CMI, UDS) or the best discretization (MAC). Hence, deterministic dependence measures such as CMI and UDS with which TCMI is most closely related are neither scale nor permutation invariant. MAC is scale invariant, but not permutation invariant. In contrast, the stochastic dependence measure MCDE is scale and permutation invariant, but only within a probabilistic tolerance (i.e., dependence scores vary between different runs of a program within a certain threshold).


\subsubsection{Baseline adjustment of TCMI}
\label{sec:experiments_baseline_adjustment}

\begin{figure*}
    \centering
    \includegraphics[width=\linewidth]{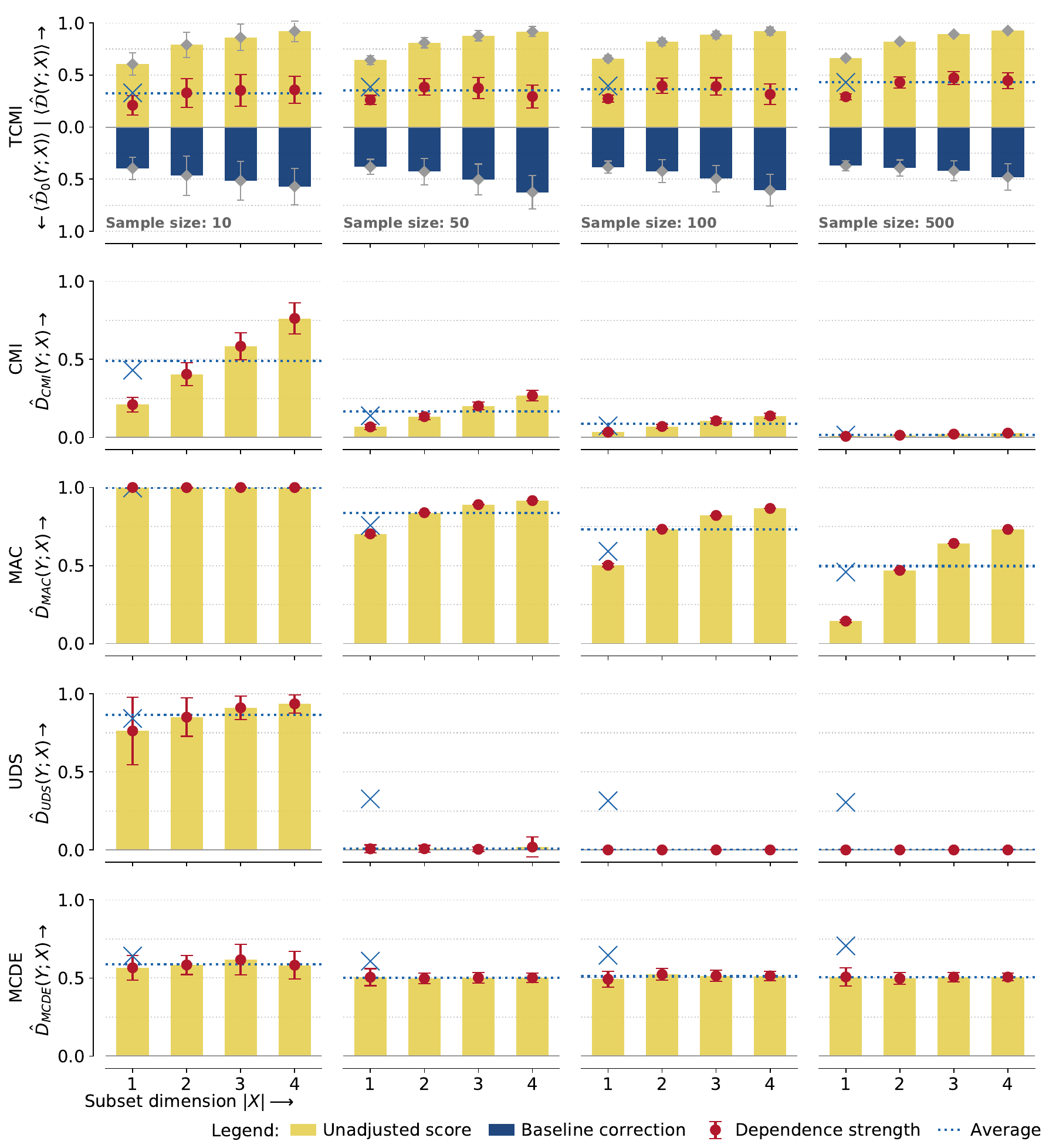}
    \caption{Fraction of cumulative information scores against increasing dimensionality for $ \{Y, \vec{X}\}$ using $10$, $50$, $100$, and $500$ data samples generated from mutually independent and uniform distributions of size $\vec{X} = \{Y, X_1, \cdots, X_4 \}$. Contributions of average fraction of total cumulative mutual information, $\langle \hat{\mathcal D}_\text{TCMI}(Y; X) \rangle$ and $\langle \hat{\mathcal D}_\text{TCMI, 0}(Y; X) \rangle$ are shown on either side of the plot and the resulting score $\langle \hat{\mathcal D}_\text{TCMI}^*(Y; X) \rangle$ as points. Error bars indicate standard deviations from repeating the experiment $50$ times. Since $X$ and $Y$ are independent, average total cumulative mutual information should be constant across subsets of features independent of sample size and subset dimensionality. While $\langle \hat{\mathcal D}_\text{TCMI}(Y; X) \rangle$ is increasing with the cardinality of the variable feature set and $\langle \hat{\mathcal D}_\text{TCMI, 0}(Y; X) \rangle$ decreasing, $\langle \hat{\mathcal D}_\text{TCMI}^*(Y; X) \rangle$ is approximately constant for a wide range of data samples $10\ldots500$ and subset dimensionality $1\ldots4$. The crosses represent the deviation of the TCMI from the constant baseline. By enlarging the feature subset with a shuffled version of the same variable, TCMI can be corrected. For comparison the dependence scores for the other investigated measures against increasing dimensionality -- cumulative mutual information (CMI: \cite{NguyenMuellerEtAl:2013}), multivariate maximal correlation analysis (MAC: \cite{NguyenMuellerEtAl:2014a}), and universal dependency analysis (UDS: \cite{NguyenMandrosEtAl:2016,WangRomanoEtAl:2017}) -- are also shown.}
    \label{fig:baseline_adjustment}
\end{figure*}

In the fourth experiment, we investigate the necessity of a baseline adjustment to estimate mutual dependences (Section~\ref{sec:baseline_adjustment}). To this end, we generated mutually independent and uniform distributions $Z = \{ Y, X_1, \ldots , X_d \}$ of dimensionality $d$ with sample sizes $10$, $50$, $100$, and $500$. We compared TCMI, CMI, MAC, UDS, and MCDE across subsets of variables of different subspace dimensionality while repeating the experiment $50$ times. Figure~\ref{fig:baseline_adjustment} summarizes the results.

By definition, the score of the dependence measures for independent random variables must be zero independent of the sample size (cf., Section~\ref{sec:baseline_adjustment}). However, none of the investigated dependence-measure scores are zero for all sample sizes. This is due to the fact that sample data are rarely exactly uniform. In practice, due to random sampling, we expect constant scores approaching $\langle \hat{\mathcal D}_\text{TCMI}^*(Y; \vec{X}) \rangle \to 0.5$ and $\hat{\mathcal D}_\text{MCDE}(Y; \vec{X}) \to 0.5$ as $n \to \infty$ independent of the dimensionality of $\vec{X} = \{ X_1, \ldots, X_d \}$ in the case if none of the variables are dependent to $Y$, and zero scores for CMI, MAC, and UDS.

Dependence scores of TCMI and MCDE are approximately constant for a wide range of data samples $10\ldots500$ and subset dimensionality $1\ldots4$ and approach $\langle \mathcal D_\text{TCMI}^* \rangle \to 0.5$ or $\hat{\mathcal D}_\text{MCDE} \to 0.5$ as $n \to \infty$ as expected. In contrast, CMI, MAC, and UDS show a clear bias towards larger dependence scores at larger subset cardinalities. Furthermore, their scores are nonzero even at larger sample sizes between mutually independent random variables. In addition, their dependence scores decrease with more data samples, indicating that these measures are unreliable in estimating the strength of mutual dependences.

In comparison to MCDE, CMI, MAC, UDS, and TCMI underestimate dependencies in the one-dimensional case when noise is present in the data. By enlarging the subset with a shuffled version of the same variable, thereby simulating a variable with noise, these measures can be corrected (Fig.~\ref{fig:baseline_adjustment}). As a result, both the corrected version of TCMI and MCDE provide a clear comparison mechanism of dependence scores across different subsets of variables, independent of the number of data samples.

\subsubsection{Bivariate normal distribution}
\label{sec:experiment_bivariate_normal_distribution}

\begin{figure}[t]
    \centering
    \includegraphics[width=\columnwidth]{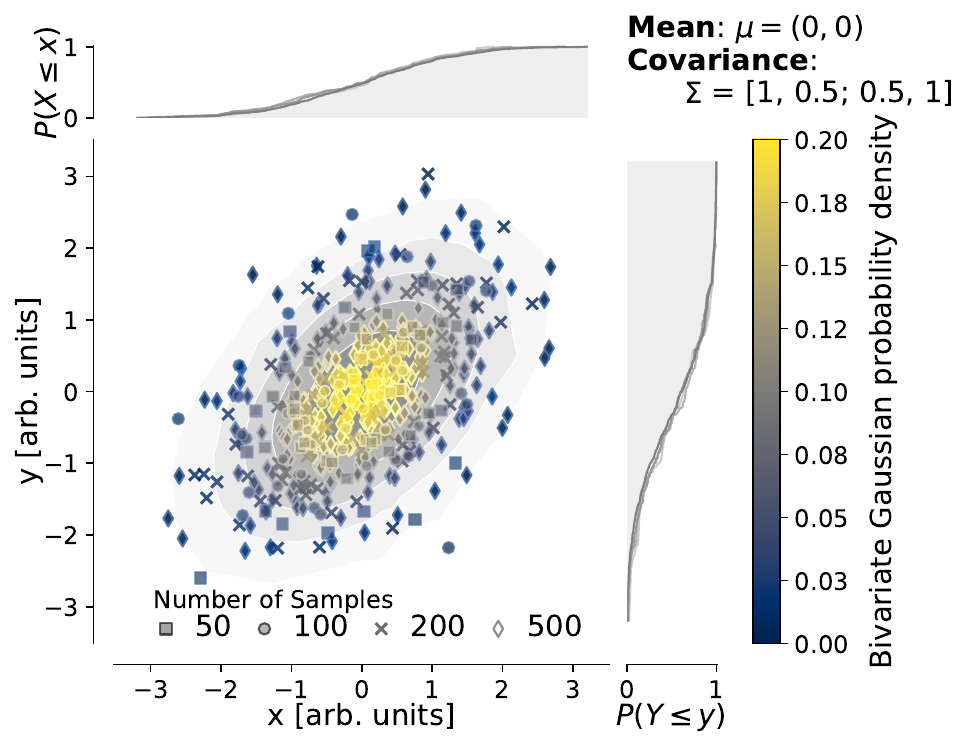}
    \caption{Bivariate normal probability distribution with mean $\mu = (0, 0)$ and covariance matrix $\Sigma = [1, 0.5; 0.5, 1]$. Shown is a scatter plot with $50$, $100$, $200$, and $500$ data samples, its cumulative probability distributions, $P(Z \leq z)$, $Z \in \{ X, Y \}$, and contour lines of equal probability densities $\in \{ 0.01, 0.02, 0.05, 0.08, 0.13 \}$.}
    \label{fig:bivariate_normal_distribution}
\end{figure}

\begin{table*}[t]
    \centering%
    \begin{tabular}{p{12em} cc @{} cc}
        \toprule
        \multirow{1}{8em}{\bfseries Dependence Measure} & \multicolumn{4}{c}{\bfseries Sample size} \\
        & 50 & 100 & 200 & 500 \\
        \midrule
        TCMI &
        \begin{minipage}[t]{0.17\textwidth}
            \{logistic,x\}=0.54 \newline \{rayleigh,weibull\}=0.53 \newline \{x,y\}=0.52 \newline \{x\}=0.35 \newline \{rayleigh\}=0.35 \newline \{laplace\}=0.35 \newline \{triangular\}=0.34
        \end{minipage} &
        \begin{minipage}[t]{0.17\textwidth}
            \{y,rayleigh\}=0.57 \newline \{y,laplace\}=0.55 \newline \{y,uniform\}=0.55 \newline \{y\}=0.46
        \end{minipage} &
        \begin{minipage}[t]{0.15\textwidth}
            \{x,y\}=0.58 \newline \{y,exponential\}=0.55 \newline \{y\}=0.48 \newline \{y,poisson\}=0.47
        \end{minipage} &
        \begin{minipage}[t]{0.18\textwidth}
            \{y,x\}=0.60 \newline \{x,normal\}=0.57 \newline \{normal,triangular\}=0.57 \newline \{x\}=0.38
        \end{minipage} \\

        \midrule
        CMI \citep{NguyenMuellerEtAl:2013} &
        \begin{minipage}[t]{0.17\textwidth}
            \{y\}=1.00 \newline \{logistic\}=1.00 \newline \{triangular\}=1.00 \newline \{laplace\}=1.00
        \end{minipage} &
        \begin{minipage}[t]{0.17\textwidth}
            \{y\}=1.00 \newline \{logistic\}=1.00 \newline \{normal\}=1.00 \newline \{triangular\}=1.00 \newline \{laplace\}=1.00
        \end{minipage} &
        \begin{minipage}[t]{0.15\textwidth}
            \{y\}=1.00 \newline \{x\}=1.00 \newline \{triangular\}=1.00 \newline \{laplace\}=1.00 \newline \{logistic\}=1.00 \newline \{normal\}=1.00
        \end{minipage} &
        \begin{minipage}[t]{0.18\textwidth}
            \{y\}=1.00 \newline \{x\}=1.00 \newline \{logistic\}=1.00 \newline \{triangular\}=1.00 \newline \{laplace\}=1.00 \newline \{normal\}=1.00
        \end{minipage} \\

        \midrule
        MAC \citep{NguyenMuellerEtAl:2014a} &
        \begin{minipage}[t]{0.17\textwidth}
            \{y,laplace\}=0.90 \newline \{y,x\}=0.89 \newline \{y,triangular\}=0.89 \newline \{y,exponential\}=0.89 \newline \{y,normal\}=0.89 \newline \{y,rayleigh\}=0.89 \newline \{y,uniform\}=0.89 \newline \{y,weibull\}=0.89 \newline \{y,logistic\}=0.89 \newline \{y\}=0.88
        \end{minipage} &
        \begin{minipage}[t]{0.17\textwidth}
            \{x,laplace\}=0.82 \newline \{x,logistic\}=0.82 \newline \{x,weibull\}=0.82 \newline \{x,triangular\}=0.82 \newline \{x,exponential\}=0.82 \newline \{x,normal\}=0.82 \newline \{x,rayleigh\}=0.82 \newline \{x,uniform\}=0.82 \newline \{x,y\}=0.82 \newline \{y\}=0.81
        \end{minipage} &
        \begin{minipage}[t]{0.15\textwidth}
            \{y\}=0.83 \newline \{x\}=0.83
        \end{minipage} &
        \begin{minipage}[t]{0.18\textwidth}
            \{y\}=0.81 \newline \{weibull\}=0.78
        \end{minipage} \\

        \midrule
        UDS \citep{NguyenMandrosEtAl:2016,WangRomanoEtAl:2017} &
        \begin{minipage}[t]{0.17\textwidth}
            \{laplace\}=0.52
        \end{minipage} &
        \begin{minipage}[t]{0.17\textwidth}
            \{y\}=0.49 \newline \{normal\}=0.48
        \end{minipage} &
        \begin{minipage}[t]{0.15\textwidth}
            \{normal\}=0.47
        \end{minipage} &
        \begin{minipage}[t]{0.18\textwidth}
            \{normal\}=0.45 \newline \{logistic\}=0.44
        \end{minipage} \\

        \midrule
        MCDE \citep{FoucheBoehm:2019,FoucheMazankiewiczEtAl:2021} &
        \begin{minipage}[t]{0.17\textwidth}
            \{y\}=0.84
        \end{minipage} &
        \begin{minipage}[t]{0.17\textwidth}
            \{x\}=0.88 \newline \{y\}=0.86
        \end{minipage} &
        \begin{minipage}[t]{0.15\textwidth}
            \{x\}=0.92 \newline \{y\}=0.88
        \end{minipage} &
        \begin{minipage}[t]{0.18\textwidth}
            \{y\}=0.94 \newline \{x\}=0.92
        \end{minipage} \\

        \bottomrule
    \end{tabular}
    \caption{Topmost feature subsets in the order of identification from the bivariate normal distribution data set with $50$, $100$, $200$, and $500$ data samples as being restricted to subset dimensionality $\leq 2$ and selected by the following dependence measures: total cumulative mutual information (TCMI), cumulative mutual information (CMI), multivariate maximal correlation analysis (MAC), universal dependency analysis (UDS), and Monte Carlo dependency estimation (MCDE).}
    \label{tab:bivariate_normal_distribution_subspaces}
\end{table*}

At last, we consider a simple feature-selection task with known ground truth, namely to find a bivariate normal distribution embedded in a high-dimensional space. For this purpose, we generated a bivariate normal distribution of size $n = 500$ from features $x$ and $y$, added additional variables such as normal, exponential, logistic, triangular, uniform, Laplace, Rayleigh, and Weibull distributions all with zero mean $\mu = 0$ and identity covariance matrix $\sigma = 1$, and augmented the feature space as described in Section~\ref{sec:limitations_of_mutual_information}. In terms of Pearson or Spearman's correlation coefficient, none of the features have coefficients of determinations higher than $1\%$ with respect to the bivariate normal distribution. Thus, without knowing the ground truth, the data set appears to be uncorrelated. However, since the ground truth is known, there are two features, namely $x$ and $y$, to completely describe the bivariate normal distribution of the data set.\\

\paragraph{Subspace search}
In order to find the two most relevant features from the high-dimensional data set, a subspace search is performed up to the second subset dimensionality. Further, feature selection is being performed for four sets of $50$, $100$, $200$, and $500$ data samples (Fig.~\ref{fig:bivariate_normal_distribution}). Results are reported in Table~\ref{tab:bivariate_normal_distribution_subspaces}.

Overall, almost all dependence measures find at least one of the two relevant features $x$, $y$, both, or at least similar distributions, such as the normal distribution. However, scores and subset sizes of relevant features decrease with larger sample sizes for MAC and UDS, while CMI identifies exact dependencies even between distributions where none dependence exists, e.g., between a Laplacian and bivariate normal distribution. In contrast, MCDE robustly finds one of the relevant features $x$ or $y$, but never finds two of them being jointly relevant. TCMI also finds one or both of the two relevant features, but scores and relevance are more determined by sample size. With sample sizes greater than 200, TCMI is the only dependence measure that correctly identifies the optimal feature subset to be $\{x,y\}$. Still, TCMI scores are lower than of the other dependence measures, even though the score increases for larger sample sizes.

\paragraph{Statistical power analysis}
To assess the robustness of dependence measures, we performed a statistical power analysis for CMI, MAC, UDS, MCDE, and TCMI and added Gaussian noise with increasing standard deviation $\sigma$ \citep{NguyenMuellerEtAl:2014a,NguyenMandrosEtAl:2016,FoucheBoehm:2019,FoucheMazankiewiczEtAl:2021}. We considered $5+1$ noise levels, distributed linearly from $0$ to $1$, inclusive. We computed the score of the bivariate normal distribution for each dependence $\Lambda=$\{CMI, MAC, UDS, MCDE, TCMI\}, i.e., $\langle \Lambda(Y; X) \rangle_\sigma$, with $n = 500$ data samples and subset $\{x,y\}$ and compared it with the score of independently drawn random data samples, $\langle \Lambda(Y; I) \rangle_0$, of the same size ($n = 500$) and dimension ($d = 1+2$). The power of a dependence measure $\Lambda$, was then evaluated as the probability $P$ of a dependence score to be larger than the $\gamma$-th percentile of the score with respect to independence $I$,
\begin{equation}
    \text{Power}_{\Lambda, \sigma}^\gamma(Y; X) := P \left ( \langle \Lambda(Y; X) \rangle_\sigma > \langle \Lambda(Y; I) \rangle_0^\gamma \right ) \ .
    \label{eq:power_dependence_measure}
\end{equation}
Essentially, the power of a dependence measure quantifies the contrast, i.e., the difference between dependence $X$ and independence $I$ at noise level $\sigma$ with $\gamma\%$ confidence. It is a relative statistical measure and depends on the strength of the dependence. Therefore, dependence strengths that are close to independence are likely to be more sensitive to noise than stronger dependences.

\begin{figure*}
    \centering
    \includegraphics[width=\linewidth]{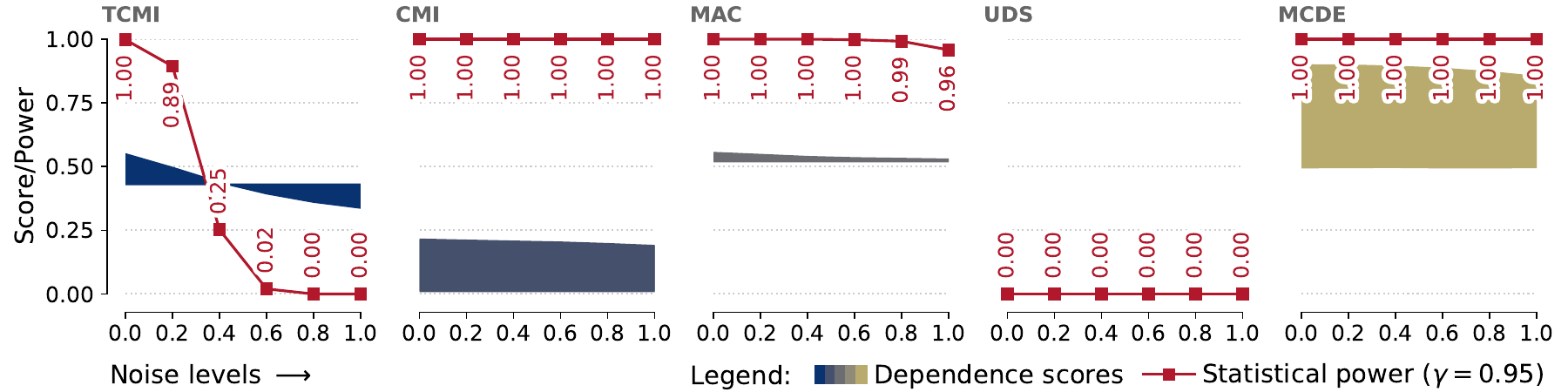}
    \caption{Statistical power analysis with $95\%$ confidence of dependence measures at different noise levels $\sigma = 0\ldots1$: total cumulative mutual information (TCMI), cumulative mutual information (CMI), multivariate maximal correlation analysis (MAC), universal dependency analysis (UDS), and Monte Carlo dependency estimation (MCDE). The diagrams also show the trends in the dependence scores of the optimal feature subset $\{x,y\}$ of the bivariate normal distribution.}
    \label{fig:statistical_power_analysis}
\end{figure*}

For our experiments, we set $\gamma = 95\%$ and repeated the experiment $500$ times. At each iteration, we  shuffled the data samples, computed the scores $\langle \Lambda(Y; X) \rangle_\sigma$ and $\langle \Lambda(Y; I) \rangle_0^\gamma$ for every dependence measure at noise level $\sigma$, and recorded the average and standard deviation of the respective dependence measures. The results of the statistical power analysis, the average score of the dependence measures and independence as well as the contrast are summarized in Figure~\ref{fig:statistical_power_analysis}.

With the exception of MAC, the statistical power of all dependence measures tends to be constant or to decrease with increasing noise level. It is remarkable that MCDE is the only dependence measure that has a high statistical power, offers a high contrast and assesses a strong dependence. In particular, the contrast of MCDE provides excellent statistics, even at noise levels much higher than TCMI. Although MAC and CMI also have high statistical power, their contrasts and dependence scores are low.

While a low contrast introduces difficulties in identifying subsets of related variables to an output, a low dependence score needs to be viewed in terms of the dependence score of all other possible subsets of features: If the subset has the highest score, it is still the subset that is most strongly related to an output given a dependence measure.

In our analysis, MCDE has the highest scores, followed by MAC, TCMI, and CMI. UDS completely fails to detect dependences in line with observations \citep{FoucheBoehm:2019,FoucheMazankiewiczEtAl:2021}. In general, TCMI is dependent on the number of samples (Eq.~\ref{eq:adjusted_fraction_of_cumulative_mutual_information}) and its contrast generally increases with more data samples. However, TCMI seems to be more sensitive and, therefore, less robust as compared to the other dependence measures. An in-depth analysis shows: the sensitivity is merely due to the moderate strength of the dependence as the statistical power is much more robust for stronger dependences in other data sets we tested.

\subsection{Case study on real-world data}
\label{sec:real_world_data}

Next, we study selected real-world data sets from KEEL and UCI Machine Learning Repository \citep{Alcala-FdezSanchezEtAl:2009,Alcala-FdezFernandezEtAl:2011,DuaGraff:2017}, and highlight TCMI for one, not restricted to, typical application of the materials-science community, namely the crystal-structure prediction of octet-binary compound semiconductors \citep{GhiringhelliVybiralEtAl:2015,GhiringhelliVybiralEtAl:2017}.

\subsubsection{KEEL and UCI regression data sets}
\label{sec:uci_regression_data sets}

We investigate how TCMI and similar dependence measures perform in real-world problems developed for multivariate regression tasks. Unfortunately, in practice, not every data set is known to have relevant features. Therefore, we compare our results with analyzed data sets with known relevant features. All in all, we consider one simulated data set from the KEEL database \citep{Alcala-FdezSanchezEtAl:2009,Alcala-FdezFernandezEtAl:2011} and two data sets from the UCI Machine Learning Repository \citep{DuaGraff:2017}:

\begin{enumerate}
    \item \textbf{Friedman \#1 regression} \citep{Friedman:1991}\\
    This data set is used for modeling computer outputs. Inputs $X_1$ to $X_5$ are independent features that are uniformly distributed over the interval $[0, 1]$. The output $Y$ is created according to the formula:
    \begin{equation}
        Y = 10 \sin (\pi X_1 X_2) + 20 (X_3 - 0.5)^2 + 10 X_4 + 5 X_5 + \epsilon \,
    \end{equation}
    where $\epsilon$ is the standard normal deviate $N(0, 1)$. In addition, the data set has five redundant variables $X_6\ldots X_{10}$ that are i.i.d\ random samples. Further, we enlarge the number of features by adding four variables $X_{11}\ldots X_{14}$ each very strongly correlated with $X_1\ldots X_4$ and generated by $f(x) = x + N(0, 0.01)$.

    \item \textbf{Concrete compressive strength} \cite{Yeh:1998} \\
    The aim of this data set is to predict the compressive strength of high performance concrete. Compressive strength is the ability of a material or structure to withstand loads that tend to reduce size. It is a highly nonlinear function of age and ingredients. These ingredients include cement, water, blast furnace slag (a by-product of iron and steel production), fly ash (a coal combustion product), superplasticizer (additive to improve the flow characteristics of concrete), coarse aggregate (e.g., crushed stone or gravel), and fine aggregate (e.g., sand).

    \item \textbf{Forest fires} \cite{CortezMorais:2007}\\
    This data set focuses on wildfires in the Montesinho Natural Park, which is located at the northern border of Portugal. It includes features such as local coordinates $x$ and $y$ where a fire occurred, the time (day, month, and year), temperature (temp), relative humidity (RH), wind, rain, and derived forest-fire features such as fine-fuel moisture code (FFMC), duff moisture code (DMC), drought code (DC), and initial spread index (ISI) to estimate the propagation speed of fire.
\end{enumerate}
For each data set, we performed feature selection using all aforementioned dependence measures (TCMI, CMI, MAC, UDS, MCDE) and compared resulting feature subsets with potentially relevant features reported from the original references. Results are summarized in Table~\ref{tab:real_world_regression_examples}.
\begin{table*}[t!]
    \centering
    \begin{tabular}{@{} ll @{}}
        \toprule
        \multirow{1}{6em}{\bfseries Dependence Measure} & \textbf{Relevant feature subsets}\\
        & (Data set, reported relevant features, feature subsets by dependence measures) \\[.5em]
        \midrule
        \multicolumn{2}{@{} p{.98\textwidth} @{}}{\textsf{Friedman \#1 regression} \citep{Friedman:1991}: $X_1\ldots X_{14}$\newline{\small Potentially relevant features: $X_1\ldots X_5$ and $X_{11}\ldots X_{14}$}\hfill[500 data samples]} \\
        \midrule
        TCMI & $\{X_{14},X_{12},X_1,X_5,X_3\}=0.79$, $\{X_{14},X_{12},X_1,X_5\}=0.77$, $\{X_{4},X_{2},X_1,X_3\}=0.77$,\\
        & $\{X_4,X_2,X_1,X_8\}=0.76$, $\{X_{14},X_{12},X_1,X_7\}=0.75$\\[.5em]

        CMI & $\{X_{14}\}=1.00$, $\{X_4\}=1.00$\\[.5em]

        MAC & \{$X_{14},X_8,X_9,X_7,X_{11},X_3,X_6,X_{10},X_{12},X_5\}=0.89$, \ldots (+ 119.981 subsets$=0.89$)\\[.5em]

        UDS & --\\[.5em]

        MCDE & $\{X_{2}\}=0.78$, $\{X_{12}\}=0.77$, $\{X_{11}\}=0.77$, $\{X_{1}\}=0.77$ \\[.5em]

        \midrule
        \multicolumn{2}{@{} p{.98\textwidth} @{}}{\textsf{Concrete compressive strength} \citep{Yeh:1998}: age, cement, water, blast furnace slag (slag), fly ash, superplasticizer (sp), coarse aggregate (coarse\_aggr), fine aggregate (fine\_aggr)\newline{\small Potentially relevant features: age, cement, water, slag}\hfill[1030 data samples]} \\
        \midrule
        TCMI & $\{\text{cement},\text{sp},\text{water},\text{coarse\_aggr},\text{fine\_aggr}\}=0.68$ \\
        & $\{\text{fine\_aggr},\text{water},\text{sp},\text{coarse\_aggr},\text{fly\_ash}\}=0.68$ \\
        & $\{\text{fine\_aggr},\text{water},\text{sp},\text{coarse\_aggr},\text{age}\}=0.68$ \\
        & $\{\text{cement},\text{coarse\_aggr},\text{water},\text{slag},\text{fine\_aggr}\}=0.68$ \\

        & $\{\text{fine\_aggr},\text{slag},\text{water},\text{coarse\_aggr},\text{age}\}=0.67$ \\
        & $\{\text{cement},\text{coarse\_aggr},\text{water},\text{sp},\text{age}\}=0.67$ \\
        & $\{\text{cement},\text{coarse\_aggr},\text{fine\_aggr},\text{sp},\text{age}\}=0.67$ \\
        & $\{\text{coarse\_aggr},\text{cement},\text{fine\_aggr},\text{water},\text{sp}\}=0.66$\\[.5em]

        CMI & $\{\text{age}\}=1.00$, $\{\text{cement}\}=1.00$, $\{\text{coarse\_aggr}\}=1.00$, $\{\text{fine\_aggr}\}=1.00$, \\
        & $\{\text{slag}\}=1.00$, $\{\text{water}\}=1.00$, $\{\text{sp}\}=0.98$\\[.5em]

        MAC & $\{\text{water},\text{coarse\_aggr},\text{fine\_aggr},\text{cement},\text{sp},\text{slag},\text{fly\_ash},\text{age}\}=0.76$\\[.5em]

        UDS & --\\[.5em]

        MCDE & $\{\text{age}\}=0.90$ \\[.5em]

        \midrule
        \multicolumn{2}{@{} p{.98\textwidth} @{}}{\textsf{Forest fires} \citep{CortezMorais:2007}: $x$, $y$, time (day, month, and year), temperature (temp), relative humidity (RH), wind, rain, fine fuel moisture code (FFMC), duff moisture code (DMC), drought code (DC), initial spread index (ISI)\newline{\small Potentially relevant features: temp, rain, RH, wind}\hfill[517 data samples]} \\
        \midrule
        TCMI & $\{\text{DMC},\text{RH},\text{ISI},\text{temp},\text{wind},\text{DC}\}=0.53$,\\
        & $\{\text{DMC},\text{RH},\text{DC},\text{temp},\text{FFMC},\text{wind}\}=0.51$ \\[.5em]

        CMI & $\{\text{temp},\text{DC}\}=1.00$, $\{\text{temp},\text{DMC}\}=1.00$, $\{\text{temp},\text{RH}\}=1.00$, \\
        & $\{\text{temp},\text{FFMC}\}=1.00$, $\{\text{FFMC},\text{DC}\}=1.00$, $\{\text{FFMC},\text{DMC}\}=1.00$, \\
        & $\{\text{FFMC},\text{RH}\}=1.00$, $\{\text{FFMC},\text{temp}\}=1.00$, $\{\text{DMC},\text{DC}\}=1.00$,\\
        &  $\{\text{DMC},\text{ISI}\}=1.00$, $\{\text{DMC},\text{RH}\}=1.00$, $\{\text{DMC},\text{month}\}=1.00$,\\
        & $\{\text{DC},\text{DMC}\}=1.00$, $\{\text{DC},\text{ISI}\}=1.00$, $\{\text{DC},\text{RH}\}=1.00$, $\{\text{DC},\text{month}\}=1.00$,\\
        & $\{\text{RH},\text{DMC}\}=1.00$, $\{\text{RH},\text{DC}\}=1.00$, $\{\text{ISI},\text{DMC}\}=1.00$, $\{\text{ISI},\text{DC}\}=1.00$,\\
        & $\{\text{temp},\text{month}\}=1.00$\\[.5em]

        MAC & $\{\text{temp},\text{RH},\text{DMC},\text{FFMC},\text{DC},\text{ISI},\text{wind},\text{day},\text{x}\}=0.85$, \\
        & $\{\text{temp},\text{RH},\text{DMC},\text{FFMC},\text{DC},\text{ISI},\text{wind},\text{day}\}=0.83$, \\
        & $\{\text{temp},\text{RH},\text{DMC},\text{FFMC},\text{DC},\text{ISI},\text{wind},\text{x}\}=0.83$\\[.5em]

        UDS & $\{\text{rain}\}=0.35$\\[.5em]

        MCDE & $\{\text{DMC},\text{temp},\text{RH}\}=0.84$, $\{\text{DMC},\text{temp},\text{DC}\}=0.82$, $\{\text{DMC},\text{temp},\text{FFMC}\}=0.81$ \\[.5em]

        \bottomrule
    \end{tabular}
    \caption{Relevant feature subsets for selected data sets from the KEEL database \citep{Alcala-FdezSanchezEtAl:2009,Alcala-FdezFernandezEtAl:2011} and UCI Machine Learning Repository \citep{DuaGraff:2017}, designed for multivariate regression tasks and feature selection as found out by total cumulative mutual information (TCMI), cumulative mutual information (CMI), multivariate maximal correlation analysis (MAC), universal dependency analysis (UDS), and Monte Carlo dependency estimation (MCDE). For comparison, potentially relevant feature subsets mentioned in the references are also included.}
    \label{tab:real_world_regression_examples}
\end{table*}

Our results show that even in the simplest example of the Friedman regression data set, two dependence measures show extreme behavior: UDS selects no variables and MAC selects all variables of the data set and therefore both do not perform any feature selection at all. Both dependence measures do not only completely fail to identify the actual dependences of the Friedman regression data set, but also fail in the concrete compressive strength, and forest fires data set. Therefore, it is likely that these dependence measures report incorrect results in other data sets and are therefore inappropriate for feature selection and dependence-assessment tasks.

CMI and MCDE partially agree with potentially relevant features from the respective references: Therefore, they may be useful when low-dimensional feature subsets need to be identified. In contrast, TCMI effectively selects all relevant variables of the Friedman regression data set. However, TCMI it is not free from selecting non-relevant variables in sub-optimal feature subsets as it reports $X_7$ or $X_8$ in the fourth or fifth best feature subset. Therefore, dependence scores need to be related with respect to the baseline adjustment term, and the lower the dependence scores are, the more likely non-relevant variables are in the subsets (cf., Section~\ref{sec:experments_interpretability_tcmi}).

Found feature subsets with TCMI for the Friedman regression data set as well as for the concrete compressive strength data set have high dependence scores. They agree well with relevant features as reported by the references, even though TCMI misses slag in the concrete compressive strength example: It is likely that variables such as fine and coarse aggregate or superplasticizer serve as a substitute for slag due to the limited number of data samples. However, we cannot test this assumption as all data samples were used to compute the dependence scores and no curated test sets are available for further tests.

In the forest-fires data set, temperature and relative humidity as well as duff moisture and drought code are not only reported by TCMI, but also by CMI and MCDE. It is therefore likely that these variables are also relevant in the forest-fires predictions, although none of them were mentioned in the reference \citep{CortezMorais:2007}. Apart from weather conditions, TCMI also includes some of the derived forest-fires variables such as duff moisture (CMD) and drought code (DC) -- these variables are indirectly related to precipitation and are used to estimate the lower and deeper moisture content of the soil. Admittedly, the TCMI scores are moderate, which indicates difficulties in assessing the mutual dependences between a set of features and the burnt area of forest fires as a whole. A detailed analysis shows that although forest fires are devastating, they are isolated events -- not enough to actually reliably identify the precursors of wildfires from the investigated data set.

\subsubsection{Octet-binary compound semiconductors}
\label{sec:octet_binary_compound_semiconductors}

Our last example is dedicated to a typical, well characterized, and canonical materials-science problem, namely the crystal-structure stability prediction of octet-binary compound semiconductors \citep{GhiringhelliVybiralEtAl:2015,GhiringhelliVybiralEtAl:2017}. Octet-binary compound semiconductors are materials consisting of two elements formed by groups of I/VII, II/VI, III/V, or IV/IV elements leading to a full valence shell. They can crystallize in rock salt (RS) or zinc blende (ZB) structures, i.e., either with ionic or covalent bindings and were already studied in the 1970's \citep{Van:1969,Phillips:1970}, followed by further studies \citep{Zunger:1980,Pettifor:1984}, and recent work using machine learning \citep{SaadGaoEtAl:2012,GhiringhelliVybiralEtAl:2015,GhiringhelliVybiralEtAl:2017,OuyangCurtaroloEtAl:2018}.

The data set consists of 82 materials with two atomic species in the unit cell. The objective is to accurately predict the energy difference $\Delta E$ between RS and ZB structures based on 8 electro-chemical atomic properties for each atomic species $A/B$ (in total 16) such as atomic ionization potential $\text{IP}$, electron affinity $\text{EA}$, the energies of the highest-occupied and lowest-unoccupied Kohn-Sham levels, $\text{H}$ and $\text{L}$, and the expectation value of the radial probability densities of the valence $s$-, $p$-, and $d$-orbitals, $r_s$, $r_p$, and $r_d$, respectively \citep{GhiringhelliVybiralEtAl:2015}. As a reference, we added Mulliken electronegativity $\text{EN} = -(\text{IP}+\text{EA})/2$ to the data set and also studied the best two features from the publication \citep{GhiringhelliVybiralEtAl:2015}
\begin{gather}
    D_1 = \frac{\text{IP}(B) - \text{EA}(B)}{r_p(A)^2} , \quad \ D_2 = \frac{\lvert r_s(A) - r_p(B) \rvert}{\exp[r_s(A)]} \ ,
    \label{eq:binary_octet_descriptors}
\end{gather}
as known dependences to show the consistency of the method as well as to probe TCMI with linearly dependent features \citep{GhiringhelliVybiralEtAl:2015}.

\begin{table*}[t!]
    \centering
    \begin{tabular}{@{} p{4em} @{}l @{\hskip3pt} rrrr @{}}
        \toprule
        \multirow{2}{5.5em}{\bfseries Subset dimension} & \multirow{2}{*}{\shortstack[l]{\bfseries Feature subsets and dependence score \\ \textbf{(TCMI)}}} & \multicolumn{4}{c}{\bfseries Metrics (GBDT)} \\
        \cmidrule{3-6}
        & & RMSE & MAE & MaxAE & $r^2$ \\

        \midrule
        \multicolumn{1}{@{}l}{6\hspace{3.5em}$\star$} & $\{D_2,\text{EA}(A),r_p(A),r_s(A),r_p(B),\text{L}(B)\}=0.84$ & 0.15 & 0.10 & 0.43 & 0.86 \\
        & $\bm{\{\text{EA}(A),r_p(A),r_s(A),\text{EN}(B),\text{L}(B),r_s(B)\}=0.82}$  & 0.12 & 0.08 & 0.32 & 0.91 \\
        & $\bm{\{\text{EA}(A),r_p(A),r_s(A),\text{EN}(B),\text{L}(B),r_p(B)\}=0.82}$  & 0.12 & 0.08 & 0.32 & 0.91 \\
        & $\bm{\{\text{EA}(A),r_p(A),r_s(A),\text{IP}(B),\text{L}(B),r_s(B)\}=0.82}$  & 0.13 & 0.09 & 0.33 & 0.90 \\
        & $\bm{\{\text{EA}(A),r_p(A),r_s(A),\text{IP}(B),\text{L}(B),r_p(B)\}=0.82}$  & 0.13 & 0.09 & 0.33 & 0.90 \\
        & $\{\text{EA}(A),r_p(A),r_s(A),\text{H}(B),\text{L}(B),r_s(B)\}=0.82$   & 0.14 & 0.10 & 0.36 & 0.87 \\
        & $\{\text{EA}(A),r_p(A),r_s(A),\text{H}(B),\text{L}(B),r_p(B)\}=0.82$   & 0.14 & 0.10 & 0.36 & 0.87 \\
        & $\{\text{EA}(A),\text{H}(A),r_d(A),r_p(A),\text{L}(B),r_d(B)\}=0.82$   & 0.15 & 0.10 & 0.45 & 0.86 \\
        & $\{\text{EA}(A),\text{H}(A),r_d(A),r_s(A),\text{L}(B),r_d(B)\}=0.82$   & 0.16 & 0.10 & 0.46 & 0.85 \\
        & $\{\text{EA}(A),\text{H}(A),r_p(A),\text{L}(B),r_d(B),r_p(B)\}=0.81$   & 0.14 & 0.10 & 0.37 & 0.88 \\
        & $\{\text{EA}(A),\text{H}(A),r_p(A),\text{L}(B),r_d(B),r_s(B)\}=0.81$   & 0.14 & 0.10 & 0.37 & 0.87 \\[.5em]

        5 & $\bm{\{\text{EA}(A),r_p(A),r_s(A),\text{IP}(B),\text{L}(B)\}=0.79}$    & 0.13 & 0.08 & 0.40 & 0.89 \\
        & $\bm{\{\text{EA}(A),r_p(A),r_s(A),\text{EN}(B),\text{L}(B)\}=0.79}$    & 0.14 & 0.08 & 0.46 & 0.88 \\
        & $\{\text{EA}(A),r_p(A),r_s(A),\text{H}(B),\text{L}(B)\}=0.79$     & 0.15 & 0.09 & 0.42 & 0.86 \\
        \multicolumn{1}{r}{$\star$} & $\{D_1,D_2,r_p(A),r_s(A),r_s(B)\}=0.79$ & 0.17 & 0.10 & 0.50 & 0.83 \\
        & $\{\text{EN}(A),r_p(A),r_s(A),\text{IP}(B),\text{L}(B)\}=0.78$    & 0.14 & 0.08 & 0.43 & 0.88 \\
        & $\{\text{EA}(A),\text{H}(A),r_p(A),\text{L}(B),r_s(B)\}=0.78$     & 0.14 & 0.10 & 0.37 & 0.88 \\
        & $\{\text{EA}(A),\text{H}(A),r_p(A),\text{L}(B),r_p(B)\}=0.78$     & 0.14 & 0.10 & 0.37 & 0.88 \\
        & $\{\text{EA}(A),\text{H}(A),r_d(A),r_p(A),\text{L}(B)\}=0.78$     & 0.17 & 0.09 & 0.51 & 0.84 \\
        & $\{\text{EA}(A),\text{H}(A),r_d(A),r_s(A),\text{L}(B)\}=0.78$     & 0.17 & 0.10 & 0.53 & 0.83 \\
        & $\{\text{EA}(A),\text{H}(A),\text{L}(A),r_s(A),\text{L}(B)\}=0.78$      & 0.18 & 0.10 & 0.55 & 0.82 \\[.5em]

        4 & $\{\text{EA}(A),r_p(A),r_s(A),\text{L}(B)\}=0.78$  & 0.16 & 0.09 & 0.49 & 0.85 \\
        & $\{\text{L}(A),r_p(A),r_s(A),r_p(B)\}=0.76$  & 0.13 & 0.09 & 0.35 & 0.90 \\
        & $\{\text{L}(A),r_p(A),r_s(A),r_s(B)\}=0.76$  & 0.13 & 0.09 & 0.33 & 0.90 \\
        & $\{\text{EN}(A),r_p(A),r_s(A),\text{L}(B)\}=0.76$  & 0.17 & 0.10 & 0.52 & 0.83 \\
        \multicolumn{1}{r}{$\star$} & $\{D1,r_p(A),r_s(A),r_s(B)\}=0.75$ & 0.15 & 0.11 & 0.37 & 0.87 \\[.5em]

        3 & $\{r_p(A),r_s(A),r_s(B)\}=0.73$   & 0.13 & 0.10 & 0.31 & 0.89 \\
        & $\bm{\{\text{IP}(A),r_p(A),\text{L}(B)\}=0.73}$    & 0.16 & 0.10 & 0.49 & 0.84 \\
        & $\{r_p(A),r_s(A),\text{L}(B)\}=0.73$    & 0.16 & 0.10 & 0.48 & 0.84 \\
        & $\bm{\{\text{EN}(A),r_p(A),\text{L}(B)\}=0.73}$    & 0.18 & 0.11 & 0.53 & 0.80 \\
        & $\{r_p(A),r_s(A),r_p(B)\}=0.72$   & 0.13 & 0.10 & 0.31 & 0.89 \\
        & $\bm{\{\text{IP}(A),r_s(A),\text{L}(B)\}=0.72}$    & 0.17 & 0.10 & 0.49 & 0.82 \\
        & $\bm{\{\text{EN}(A),r_s(A),\text{L}(B)\}=0.72}$    & 0.18 & 0.11 & 0.52 & 0.80 \\

        \multicolumn{1}{r}{$\star$} & $\{D_1,r_s(A),r_p(B)\}=0.70$ & 0.15 & 0.11 & 0.40 & 0.86 \\[.5em]

        \multicolumn{1}{@{}l}{2\hspace{3.5em}$\star$} & $\{D_1,r_s(B)\}=0.71$ & 0.19 & 0.14 & 0.52 & 0.76 \\
        & $\{r_s(A),\text{L}(B)\}=0.69$ & 0.18 & 0.12 & 0.49 & 0.80 \\
        & $\{r_s(A),r_s(B)\}=0.67$ & 0.14 & 0.10 & 0.34 & 0.88 \\
        \multicolumn{1}{r}{$\star$} & $\{D_1,D_2\}=0.62$ & 0.19 & 0.14 & 0.53 & 0.77 \\[.5em]

        \multicolumn{1}{@{}l}{1\hspace{3.5em}$\star$} & $\{D_1\}=0.57$ & 0.23 & 0.18 & 0.56 & 0.69 \\
        & $\{r_s(A)\}=0.56$ & 0.21 & 0.15 & 0.53 & 0.75 \\
        & $\{r_p(A)\}=0.55$ & 0.21 & 0.15 & 0.54 & 0.75 \\

        \midrule
        \multicolumn{2}{@{}l}{All 16 features (GBDT reference):} & 0.15 & 0.09 & 0.45 & 0.86 \\
        \bottomrule
    \end{tabular}
    \caption{Relevant feature subsets for the octet-binary compound semiconductors data set as found out by total cumulative mutual information (TCMI) showing the most relevant feature subsets of each cardinality. For comparison, best feature subsets for $D_1 = D_1(\text{IP}(B),\text{EA}(B), r_p(A))$ and $D_2 = D_2(r_s(A), r_p(B))$ from reference \citep{GhiringhelliVybiralEtAl:2015} (entries with a star $\star$) are also listed. Bold feature subsets mark subsets with interchangeable variables $\text{EN}$ and $\text{IP}$. The table also shows statistics of constructed machine-learning models using the gradient boosting decision tree (GBDT) algorithm \citep{Friedman:2001} with 10-fold cross-validation: root-mean-squared error (RMSE), mean absolute error (MAE), maximum absolute error (MaxAE), and Pearson coefficient of determination ($r^2$). Units are in electronvolts (eV).}
    \label{tab:binary_octets_subsets}
\end{table*}

To predict the energy difference $\Delta E$ between RS and ZB structures, we performed a subspace search with TCMI to identify the subset of features that exhibit the strongest dependence on $\Delta E$. Results are summarized in Table~\ref{tab:binary_octets_subsets}. In total, the strongest dependence on $\Delta E$ was found with six features from both atomic species, $A$ and $B$, before TCMI decreased again with seven features.

Results reveal that there are several feature subsets that are found to be optimal among different cardinalities. We note that TCMI never selects Mulliken electronegativity $\text{EN}$ together with either electron affinity $\text{EA}$ or ionization potential $\text{IP}$ for the same atomic species. We also note that $\text{EN}$ can be replaced by $\text{IP}$ (see bold feature subsets in Tab.~\ref{tab:binary_octets_subsets}). However, $\text{EN}$ cannot be replaced by $\text{EA}$, as $\text{EN}$ is found to be stronger linearly correlated with $\text{IP}$ than with $\text{EA}$ and hence results in slightly smaller TCMI values (by at least 0.02 in case of the optimal subsets, not shown in the table). Results therefore do not only corroborate the functional relationship between $\text{EN}$, $\text{IP}$, and $\text{EA}$, but also the consistency of TCMI.

Furthermore, TCMI indicates that features, like the atomic radii $r_s(B)$ and $r_p(B)$ or the energies $\text{EN}(B)$, $\text{H}(B)$, $\text{H}(B)$ and $\text{IP}(B)$ of IV to VIII elements, can be used interchangeably without reducing the dependence scores. Indeed, by assessing dependences between pairwise feature combinations, TCMI identifies $r_s(B)$ and $r_p(B)$ to be strongly dependent and $\text{EN}(B)$, $\text{H}(B)$, and $\text{IP}(B)$ strongly dependent, consistent with bivariate correlation measures such as Pearson or Spearman. In numbers, the Pearson coefficient of determination ($r^2$) between the atomic radii $r_s$ and $r_p$ are $r^2(r_s(A), r_p(A)) = 0.94$, $r^2(r_s(B), r_p(B)) = 0.99$ and the Pearson coefficient of determination between Mulliken electronegativity and ionization potential or electron affinity is $r^2(\text{EN}(B), \text{IP}(B)) = 0.96$, or $r^2(\text{EN}(B), \text{H}(B)) = 0.99$, respectively. These findings illustrate that TCMI assigns similar scores to collinear features.

\enlargethispage{-1em}
Features $D_1$ and $D_2$ (Eq.~\ref{eq:binary_octet_descriptors}) from the reference \citep{GhiringhelliVybiralEtAl:2015}, are combinations of atomic properties that best represent $\Delta E$ linearly,
\begin{align}
    D_1 & = D_1(\text{IP}(B),\text{EA}(B), r_p(A)) \ ,\\
    D_2 & = D_2(r_s(A), r_p(B)) \ .
\end{align}
\vspace{1em}

\noindent As such, they incorporate knowledge that generally lead to higher TCMI scores for the same feature subset cardinality. While this applies to the first and second subset dimensions, feature subsets with the aforementioned features $D_1$, $D_2$ are on par with feature subsets based on atomic properties at higher dimensions. However, $D_1$ and $D_2$ are not selected consistently by TCMI because TCMI does not make any assumption about the linearity of the dependency $(D_1, D_2) \mapsto \Delta E$. This suggests that the linear combination of $D_1$ and $D_2$ is a good, but not complete, description of the energy difference $\Delta E$.

\begin{figure*}[t]
    \centering
    \includegraphics[width=\linewidth]{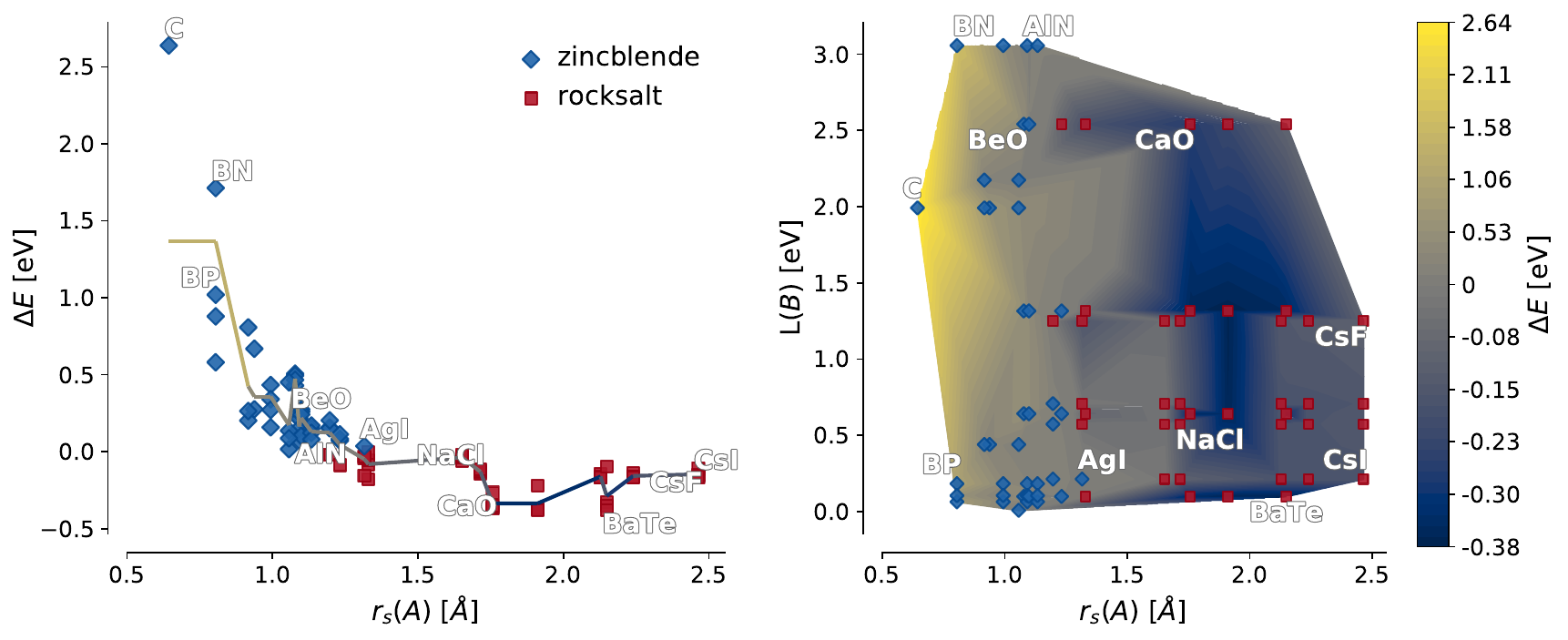}
    \caption{Feature spaces of the topmost selected feature subsets for one (left) and two dimensions (right). Shown are the two classes of crystal-lattice structures as diamonds (zinc blende) and squares (rock salt), their distribution, and the trend line/manifold in the prediction of the energy difference $\Delta E$ between rock salt and zinc blende. The trend line/manifold was computed from with the gradient boosting decision tree algorithm \citep{Friedman:2001} and 10-fold cross validation. For reference, some octet-binary compound semiconductors are labeled.}
    \label{fig:binary_octet_model}
\end{figure*}

\enlargethispage{-1em}
A visualization of relevant subsets also reveals clear monotonous relationships in one and two dimensions (Fig.~\ref{fig:binary_octet_model}). In addition, we constructed machine-learning models for each feature subset and report model statistics for the prediction of $\Delta E$ along with statistics of the full feature set (Tab.~\ref{tab:binary_octets_subsets}). The details can be found in the appendix. We partitioned the data set into $k = 10$ groups (so-called folds) and generated $k$ machine-learning models, using 9 folds to generate the model and the $k$-th fold to test the model (10-fold cross validation). To reduce variability, we performed five rounds of cross-validation with different partitions and averaged the rounds to obtain an estimate of the model's predictive performance. For the machine-learning models we used the gradient boosting decision tree algorithm (GBDT) \citep{Friedman:2001}. GBDT is resilient to feature scaling (Eq.~\ref{eq:positive_monotonic_transformations}) just like TCMI and is one of the best available, award-winning, and versatile machine-learning algorithm for classification and regression \citep{NatekinKnoll:2013,Fernandez-DelgadoCernadasEtAl:2014,CouronneProbstEtAl:2018}. Notwithstanding this, traditional methods sensitive to feature scaling may show superior performances for data sets with sample sizes larger than the number of considered features \citep{LuPetkova:2014} (compare also model performances in Tab.~\ref{tab:binary_octets_subsets} with references \citep{GhiringhelliVybiralEtAl:2015,OuyangCurtaroloEtAl:2018,GhiringhelliVybiralEtAl:2017}).

Machine-learning models are designed to improve with more data and a feature subset that best represents the data for the machine-learning algorithm \citep{Friedman:2001,JamesWittenEtAl:2013}. Therefore, we expect a general trend of higher model performances with larger feature-subset cardinalities. Furthermore, we do not expect that the optimal feature subset of TCMI performs best for every machine-learning model (``No free lunch'' theorem: \cite{Wolpert:1996a,Wolpert:1996b,WolpertMacready:1995,WolpertMacready:1997}) as an optimal feature subset identified by the feature-selection criterion TCMI may not be same according to other evaluation criteria such as root-mean-squared error (RMSE), mean absolute error (MAE), maximum absolute error (MaxAE), or Pearson coefficient of determination ($r^2$). This fact is evident in our analysis. The choice of GBDT may not be optimal because its predictive performance generally decreases with the number of variables (compare the model performance with all 16 variables to a subset with two or four variables, Tab.~\ref{tab:binary_octets_subsets}). However, to the best of our knowledge, there is no other machine-learning algorithm that models data without making assumptions about the functional form of dependence, is independent of an intrinsic metric, and can operate on a small number of data samples. Therefore, we focus only on the predictive performance of the found subsets compared to the predictive performance of the identified features with respect to all variables in the data set (Tab.~\ref{tab:binary_octets_subsets}).

Results confirm the general trend of higher model performances with larger feature-subset cardinalities and show that the initial subset of 16 variables can be reduced down to 6 variables without decreasing model performances. Essentially, feature subsets with three to four variables are already as good as a machine-learning model with all 16 variables, where the large number of variables already start to degrade the prediction performance of the GBDT model. The overall performance gradually increases with the subset cardinality. However, our analysis identifies significant variability in performance with a higher standard deviation for feature subsets at smaller dependence scores than for larger values.

An exhaustive search for the best GBDT model yields an optimum of seven features to best predict the energy difference between rock salt and zinc blende crystal structures with $D1$ and $D2$ neglected,
\begin{gather*}
    \{\text{EA}(A),\text{IP}(A),r_d(A),r_p(A),\text{IP}(B),r_s(B),r_p(B)\} \\
    \text{RMSE}: 0.11,\ \text{MAE}: 0.08,\ \text{MaxAE}: 0.27,\ r^2: 0.92 \ .
\end{gather*}
In contrast to the optimal feature subsets of TCMI (cf., Tab.~\ref{tab:binary_octets_subsets}), the optimal GBDT feature set is a variation of optimal feature subsets of TCMI with highest-occupied Kohn-Sham level and ionization potential interchanged, $\text{H}(A) \leftrightarrow \text{IP}(A)$, and lowest-unoccupied Kohn-Sham level, $\text{L}(B)$, missing. Model performances demonstrate that the optimal feature subsets of TCMI are close to the model's optimum and corroborate the usefulness of TCMI in finding relevant feature subsets for machine-learning predictions. Slight differences in performances are mainly due to the variances of the cross-validation procedure and the small number of 82 data samples, which effectively limited the reliable identification of larger feature subsets in the case of TCMI (Tab.~\ref{tab:bivariate_normal_distribution_subspaces}).

\section{Discussion}
\label{sec:discussion}

Although TCMI is a non-parametric, robust, and deterministic measure, the biggest limitation is its computational complexity. For small data sets ($n < 500$) and feature subsets ($d < 5$) feature selection finishes in minutes to hours on a modern computer. For larger data sets, however, TCMI scales with $\mathcal O(n^d)$ and quickly exceeds any realizable runtime. Furthermore, the search for the optimal feature subset also needs to be improved. Even though in our analysis only a fraction of less than one percent of the possible search space had to be evaluated, TCMI was evaluated hundreds of thousands of times. Future research towards pairwise evaluations \citep{PengLongEtAl:2005}, Monte Carlo sampling \citep{FoucheBoehm:2019,FoucheMazankiewiczEtAl:2021}, or gradual evaluation of features based on iterative refinement strategies of sampling will show to what extent the computational costs of TCMI can be reduced.

A further limitation is that non-relevant variables may be selected in the optimal feature subsets, when only a limited amount of data points is available (cf., Section~\ref{sec:experiment_bivariate_normal_distribution}). By construction, the identification of feature subsets is dependent on the feature-selection search strategy (cf., Section~\ref{sec:introduction}). The results show that it is critical to use optimal search strategies because sub-optimal search strategies can report subsets of features that are not related to an output. Even if the exhaustive search for feature subsets is computationally intensive, it can be implemented efficiently, e.g., by using the branch-and-bound algorithm. In our implementation, the branch-and-bound algorithm was used to search for optimal, i.e., minimal non-redundant feature subsets. However, as our results demonstrate, different feature subsets with few or no common features may lead to similar dependence scores. The main rationale for this outcome is that the features may be correlated with each other and therefore contain redundant information about dependences. Including these redundant features will surely lead to a higher stability of the method, more consistent results, and better insights into the actual dependence. If a machine-learning algorithm is given, the best option at present is to generate predictive models for each of the found feature subsets and select the one that works best.

\section{Conclusions}
\label{sec:conclusions}

We constructed a non-parametric and deterministic dependence measure based on cumulative probability distribution \citep{RaoChenEtAl:2004,Rao:2005} to propose fraction of cumulative mutual information $\mathcal D(Y; \vec{X})$, an information-theoretic divergence measure to quantify dependences of multivariate continuous distributions. Our measure can be directly estimated from sample data using well-defined empirical estimates (Section~\ref{sec:theoretical_background}). Fraction of cumulative mutual information quantifies dependences without breaking permutation invariance of feature exchanges, i.e., $\mathcal D(Y; \vec{X}) = \mathcal D(Y; \vec{X}')$ for all $\vec{X}' \in \operatorname{perm}(\vec{X})$, while being invariant under positive monotonic transformations. Measures based on mutual information are monotonously increasing with respect to the cardinality of feature subsets and sample size. To turn fraction of cumulative mutual information into a convex measure, we related the strength of a dependence with the dependence of the same set of variables under the independence assumption of random variables \citep{VinhEppsEtAl:2009,VinhEppsEtAl:2010}. We further constructed a measure based on residual cumulative probability distributions and introduced total cumulative mutual information $\langle \hat{\mathcal D}_\text{TCMI}^*(Y; \vec{X}) \rangle$.

Tests with simulated and real data corroborate that total cumulative mutual information is capable of identifying relevant features of linear and nonlinear dependences. The main application of total cumulative mutual information is to assess dependences, to reduce an initial set of variables before processing scientific data, and to identify relevant subsets of variables, which jointly have the largest mutual dependence and minimum redundancy with respect to an output. The performance of the total cumulative mutual information is still exponential and thus outweighs potential benefits of TCMI. In future works, we will address the performance issues of TCMI, the stability of identified feature subsets, and provide a feature-selection framework that is also suitable for discrete, continuous, and mixed data types. We will also apply TCMI to current problems in the physical sciences with a practical focus on the identification of feature subsets to simplify subsequent data-analysis tasks.

Since total cumulative mutual information identifies dependences with strong mutual contributions, it is applicable to a wide range of problems directly operating on multivariate continuous data distributions. In particular, it does not need to quantize variables by using probability density estimation, clustering, or discretization prior to estimating the mutual dependence between variables. Thus, total cumulative mutual information has the potential to promote an information-theoretic understanding of functional dependences in different research areas and to gain more insights from data.

\begin{acknowledgements}
    This research received funding from the European Research Council (ERC) under the European Unions Horizon 2020 research and innovation program (Grant Agreement No. 676580: the NOMAD Laboratory and European Center of Excellence and Grant Agreement No. 740233: TEC1p) and from BiGmax, the Max Planck Society's Research Network on Big-Data-Driven Materials Science. B.R. acknowledges financial support from the Max Planck Society. L.M.G. acknowledges support from Berlin Big-Data Center (Grant Agreement No. 01IS14013E). The authors thank J. Vreeken, M. Boley and P. Mandros for inspiring discussions and for carefully reading the manuscript.
\end{acknowledgements}

\section*{Supplementary information}

We implemented total cumulative mutual information in Python. Our Python-based implementation is part of B.R.'s doctoral thesis and is made publicly available under a \href{https://opensource.org/licenses/Apache-2.0}{Apache License 2.0}.

All data and scripts involved in producing the results can be downloaded from Zenodo (\url{https://doi.org/10.5281/zenodo.6577261}). An online tutorial to reproduce the main results presented in this work can also be found on GitHub (\url{https://github.com/benjaminregler/tcmi}) or in the NOMAD Analytics Toolkit (\url{https://analytics-toolkit.nomad-coe.eu/public/user-redirect/notebooks/tutorials/tcmi.ipynb}).

\section*{Conflict of interest}

The authors declare that they have no conflict of interest.

\section*{Corresponding authors}

Correspondence to \href{mailto:regler@fhi-berlin.mpg.de}{B. Regler (regler@fhi-berlin.mpg.de)} or \href{mailto:ghiringhelli@fhi-berlin.mpg.de}{L. M. Ghiringhelli (ghiringhelli@fhi-berlin.mpg.de)}.

\appendix\normalsize

\section*{Appendix}

\section{Baseline adjustment}

Dependence measurements that assign stronger dependences for larger subsets of features independently of the underlying relationship are considered biased \citep{VinhEppsEtAl:2009}. To actually compare dependence measures across subsets of variables and different cardinality, dependence measures need to be adjusted. Baseline adjustment is addressed by eliminating the inherent bias of the measure, so that the dependence measure becomes constant under the independence assumption of random variables. The baseline adjustment was discussed for mutual information in \citep{VinhEppsEtAl:2009,VinhEppsEtAl:2010,RomanoBaileyEtAl:2014,MandrosBoleyEtAl:2017}. Following the notation of \cite{VinhEppsEtAl:2009}, we derive the baseline adjustment for cumulative mutual information.

A common model of randomness is the hypergeometric model (also called permutation model)  \citep{Lancaster:1969,VinhEppsEtAl:2009,RomanoBaileyEtAl:2014}. It uniformly and randomly generates $m$ distinct permutations of pairs $M$ with probability $\mathcal P(Y; X \vert M)$ by permuting all values of each variable in the data set,
\begin{equation}
    \hat{\mathcal I}_0(Y; X) = \sum_{\substack{M \in \mathcal M}} \hat{\mathcal I}(Y; X \vert  M) \mathcal P(Y; X \vert  M) \ .
\end{equation}
The baseline-adjusted cumulative fraction of the information can be obtained by subtracting fraction of the cumulative information (Eq.~\ref{eq:cumulative_mutual_information}) from the expected fraction of the cumulative information under the assumption of independent and identical distributed random variables,
\begin{gather}
    \hat{\mathcal I}^*(Y; X) = \hat{\mathcal I}(Y; X) - \hat{\mathcal I}_0(Y; X) \ , \\
    \hat{\mathcal D}^*(Y; X) = \hat{\mathcal D}(Y; X) - \hat{\mathcal D}_0(Y; X) = \frac{\hat{\mathcal I}^*(Y; X)}{\hat{\mathcal H}(Y)} \ .
\end{gather}%
\begin{figure}[t]
    \centering
    \begin{tabular}{c|lllll|l|}
        \multicolumn{1}{c}{$\bf Y \backslash X$} & $\tilde{X}_1$ & $\cdots$ & $\tilde{X}_j$ & $\cdots$ & \multicolumn{2}{l}{$\tilde{X}_c$} \\
        \cline{2-7}
        $\tilde{Y}_1$ & $n_{11}$ & $\cdots$ & $\cdot$  & $\cdots$ & $n_{1c}$ & $a_1$    \\
        $\vdots$      & $\vdots$ & ~        & $\vdots$ & ~        & $\vdots$ & $\vdots$ \\
        $\tilde{Y}_i$ & $\cdot$  & ~        & $n_{ij}$ & ~        & $\cdot$  & $a_i$    \\
        $\vdots$      & $\vdots$ & ~        & $\vdots$ & ~        & $\vdots$ & $\vdots$ \\
        $\tilde{Y}_r$ & $n_{r1}$ & $\cdots$ & $\cdot$  & $\cdots$ & $n_{rc}$ & $a_r$    \\
        \cline{2-7}
        \multicolumn{1}{c|}{} & $b_1$ & $\cdots$ & $b_j$ & $\cdots$ & $b_c$ & \multicolumn{1}{|c}{} \\
        \cline{2-6}
    \end{tabular}
    \caption{A $r \times c$ cumulative contingency table $\mathcal{M}$ related to two clusterings $\tilde{X}$ and $\tilde{Y}$ with row marginals, $a_i = \sum_{j = 1}^c n_{ij}$, and column marginals, $b_j = \sum_{i =1}^r n_{ij}$. The two marginal sum vectors $a = [a_i]$ and $b = [b_j]$ are constant and satisfy the fixed marginals condition, $\sum_{i = 1}^{r} a_i = \sum_{j = 1}^{c} b_j = N$.}
    \label{tab:contingency_table}
\end{figure}%
Specifically, the average cumulative mutual information between all different permutations with $\lvert X_i \rvert = a_i$, $i = 1, \cdots, r$ and $\lvert Y_j \rvert = b_j$, $j = 1, \cdots, c$ has constant marginal sum vectors $a =[a_i]$ and $b =[b_j]$. Therefore, the cumulative information overlap between $X$ and $Y$,
\begin{equation}
    \begin{aligned}
        \hat{\mathcal I}_0(Y; X \vert  M) & = \hat{\mathcal I}_0(a, b \vert  M = [n_{ij}]_{j = 1 \cdots c}^{i = 1 \cdots r}) \\
        & = - \sum_{i = 1}^{r - 1} \sum_{j = 1}^{c} \Delta y_i(M) \frac{n_{ij}}{n} \log \frac{n_{ij}}{b_j} \ ,
    \end{aligned}
    \label{eq:correction_for_chance}
\end{equation}
can be summarized in the form of a $r \times c$ cumulative contingency table, $M = [n_{ij}]_{j = 1\cdots c}^{i = 1 \cdots r}$ (Fig.~\ref{tab:contingency_table}), with $n_{ij}$ as being a specific realization of the joint cumulative probability given row marginal $a_i$ and column marginal $b_j$.

By rearranging the sums in Eq.~\ref{eq:correction_for_chance} and writing the sum over the entire permutation of variable values as a sum over all permutations of possible values of $n_{ij}$, we get
\begin{align*}
    \hat{\mathcal I}_0(Y; X) & = - \sum_{M \in \mathcal{M}} \sum_{i = 1}^{r - 1} \sum_{j = 1}^{c} \Delta y_i(M) \frac{n_{ij}}{n} \log \frac{n_{ij}}{b_j} \mathcal P(Y; X \vert  M)\\
    & = - \sum_{i = 1}^{r - 1} \sum_{j = 1}^{c} \sum_{n_{ij}} \Delta y_i(n_{ij}, a_i, b_j \vert  M ) \\
    & \qquad \cdot \frac{n_{ij}}{n} \log \frac{n_{ij}}{b_j} \mathcal P(n_{ij}, a_i, b_j \vert  M) \ ,
\end{align*}
where $\mathcal P(n_{ij}, a_i, b_j \vert  M)$ is the probability to encounter an associative cumulative contingency table subject to fixed marginals.

The probability to encounter an associative cumulative contingency table subject to fixed marginals, with the cell at the $i$-th row and $j$-th column equals to $n_{ij}$, is given by the hypergeometric distribution,
\begin{align}
    \mathcal P(n_{ij}, a_i, b_j \vert  M) & = \mathcal P(b_j - n_{ij}, r - 1, r - i, b_j - 1) \notag \\
    & = \left . \binom{r - i}{b_j - n_{ij}} \binom{i - 1}{n_{ij} - 1} \middle / \binom{r - 1}{b_j - 1} \right . \ .
\end{align}
The hypergeometric distribution describes the probability of $b_j - n_{ij}$ successes in $b_j - 1$ draws without replacement where the finite population consists of $r - 1$ elements, of which $r - i$ are classified as successes. It is limited by the number of successes that must not exceed the limit of $\max(0, i + b_j - r) \leq n_{ij} \leq \min(i, b_j)$.
Similar, the distance $\Delta y_i(M)$ between two consecutive ordered values is described by a binomial distribution,
\begin{equation}
    \Delta y_i(n_{ij}, a_i, b_j \vert  M) = \frac{1}{\mathcal N} \sum_{k = 1}^{k_\text{max}} \binom{r - k - 1}{b_j - 2} \bigl ( y_{(i + k)} - y_{(i)} \bigr ) \ ,
\end{equation}
where the upper limit is given by $k_\text{max} = \min(n - b_j + 1, r - i)$ and $\mathcal N$ is the normalization constant:
\begin{equation}
    \mathcal N = \sum_{k = 1}^{k_\text{max}} \binom{r - k - 1}{b_j - 2} \ . 
\end{equation}
Summarizing all the single parts of Eq.~\ref{eq:correction_for_chance}, the final formula for the expected fraction of cumulative information under the assumption of the hypergeometric model of randomness is given by
\begin{multline}
    \hat{\mathcal I}_0(Y; X) = - \sum_{i = 1}^{r - 1} \sum_{j = 1}^{c} \sum_{n_{ij}} \Delta y_i(M \vert  n_{ij}, a_i, b_j) \frac{n_{ij}}{n} \log \left ( \frac{n_{ij}}{b_j} \right ) \\
    \cdot \frac{(r - i)! (i - 1)! (b_j - 1)! (r - b_j)!}{(b_j - n_{ij})! (r - i - b_j + n_{ij})! (n_{ij} - 1)! (i - n_{ij})! (r - 1)!} \ .
    \label{eq:expected_mutual_information}
\end{multline}

\section{Monotonicity conditions for total cumulative mutual information}

In the following we will prove that expected cumulative mutual information under the independence assumption of random variables $\hat{\mathcal I}_0(Y; X)$ is monotonically increasing with respect to the number of features in the subset, i.e.,
\begin{equation}
    \hat{\mathcal I}_0(Y; X) \leq \hat{\mathcal I}_0(Y; X') \text{ for } X \subset X' \subseteq \vec{X}
\end{equation}
with $X' = X \cup \{ \chi \}$ and some $\chi \not\in X$. For reference, we will closely follow the proof for the baseline correction term in the discrete case with mutual information \citep{MandrosBoleyEtAl:2017}.

Let the row and column marginals of $Y, X, X'$ be $a_i$ for $i = 1\ldots R$, $b_j$ for $j = 1\ldots C$ and $b_j'$ for $j = 1\ldots C'$, respectively. We note that $C' > C$. In order to show that
\begin{multline}
    \sum_{\substack{M \in \mathcal M}} \hat{\mathcal I}(Y; X \vert  M) \mathcal P(Y; X \vert  M) \\
    \leq \sum_{\substack{M' \in \mathcal M'}} \hat{\mathcal I}(Y; X \vert  M') \mathcal P(Y; X \vert  M') \ .
\end{multline}
we define a relation between the cumulative contingency tables $\mathcal{M} = \mathcal{M}(Y; X)$ and $\mathcal{M'} = \mathcal{M}(Y; X')$ via the projection operator $\pi: \mathcal{M'} \to \mathcal{M}$. The projection operator links the projection $\pi: V(X') \to V(X)$ of values from $X'$ to values of $X$ defined by $\pi(X') = X$ with the projection to the sets of cumulative contingency tables by finding the counts in the column corresponding to $X \in V(X)$ of $\pi(M')$ as the sum of the columns in $M'$ corresponding to $\pi^{-1}(X)$. Therefore, it remains to show that for all $M \in \mathcal{M}$ holds:
\begin{multline}
    \hat{\mathcal I}(Y; X \vert  M) \mathcal P(Y; X \vert  M) \\
    \leq \sum_{M' \in \pi(M)} \hat{\mathcal I}(Y; X \vert  M') \mathcal P(Y; X \vert  M') \ .
\end{multline}
From the chain rule of cumulative mutual information \citep{RaoChenEtAl:2004,WangVemuriEtAl:2003,Rao:2005}, it follows that $\hat{\mathcal I}(Y; X \vert  M) \leq \hat{\mathcal I}(Y; X \vert  M')$ for $M = \pi(M')$. Thus, showing the relation $\mathcal P(Y; X \vert  M) = \sum_{M' \in \pi(M)} \mathcal P(Y; X \vert  M')$ concludes the proof. We will show the proof by contradiction.

\begin{table*}[t!]
    \centering
    \begin{tabular}{p{2em} @{}l @{\hskip3pt} rrrr}
        \toprule
        \multirow{2}{4em}{\bfseries Subset dimension} & {\bfseries \newline Feature subsets and dependence score} & \multicolumn{4}{c}{\bfseries Metrics (GBDT)} \\
        \cmidrule{3-6}
        & & RMSE & MAE & MaxAE & $r^2$ \\
        \midrule

        \multicolumn{6}{l}{\itshape Cumulative mutual information (CMI):} \\
        2 & $\{\text{IP}(A),L(A)\} = 1.00$ & 0.19 & 0.12 & 0.57 & 0.79 \\
        & $\{\text{IP}(A),r_d(A)\} = 1.00$ & 0.20 & 0.13 & 0.58 & 0.77 \\
        & $\{\text{H}(A),r_d(A)\} = 1.00$ & 0.21 & 0.14 & 0.61 & 0.75 \\[.5em]

        1 & $\{r_p(A)\} = 0.24$ & 0.21 & 0.15 & 0.54 & 0.75 \\[.5em]

        \multicolumn{6}{l}{\itshape Multivariate maximal correlation analysis (MAC):} \\
        4 & $\{\text{EA}(A),H(A),r_p(A),r_s(A)\} = 0.99$ & 0.18 & 0.12 & 0.50 & 0.83 \\
        & $\{\text{EA}(A),IP(A),r_p(A),r_s(A)\} = 0.99$ & 0.18 & 0.12 & 0.50 & 0.83 \\
        & $\{\text{H}(A),IP(A),r_p(A),r_s(A)\} = 0.99$ & 0.18 & 0.12 & 0.50 & 0.83 \\[.5em]

        3 & $\{\text{EA}(A),r_p(A),r_s(A)\} = 0.98$ & 0.18 & 0.12 & 0.50 & 0.83 \\
        & $\{\text{EA}(A),\text{H}(A),r_s(A)\} = 0.98$ & 0.18 & 0.51 & 0.49 & 0.82 \\
        & $\{\text{EA}(A),\text{IP}(A),r_s(A)\}=0.98$ & 0.18 & 0.12 & 0.50 & 0.83 \\[.5em]

        2 & $\{\text{EA}(A),r_s(A)\} = 0.97$ & 0.18 & 0.12 & 0.51 & 0.82 \\
        & $\{\text{H}(A),r_s(A)\} = 0.97$ & 0.18 & 0.12 & 0.51 & 0.82 \\
        & $\{\text{IP}(A),r_s(A)\} = 0.97$ & 0.18 & 0.12 & 0.50 & 0.83 \\[.5em]

        1 & $\{r_s(A)\} = 0.88$ & 0.21 & 0.15 & 0.55 & 0.75 \\
        & $\{r_p(A)\} = 0.84$ & 0.21 & 0.15 & 0.54 & 0.75 \\
        & $\{\text{L}(A)\} = 0.88$ & 0.25 & 0.18 & 0.65 & 0.63 \\[.5em]

        \multicolumn{6}{l}{\itshape Universal dependency analysis (UDS):} \\
        9 & $\{\text{EA}(A),\text{EN}(A),\text{H}(A),\text{L}(A),r_d(A),r_p(A),r_s(A),\text{EA}(B),\text{L(B)}\} = 0.87$ & 0.20 & 0.11 & 0.62 & 0.78 \\[.5em]

        8 & $\{\text{EA}(A),\text{EN}(A),\text{H}(A),\text{L}(A),r_d(A),r_p(A),\text{EA}(B),\text{L}(B)\} = 0.86$ & 0.20 & 0.11 & 0.62 & 0.78 \\
        & $\{\text{EA}(A),\text{EN}(A),\text{H}(A),\text{L}(A),r_d(A),r_s(A),\text{EA}(B),\text{L}(B)\} = 0.86$ & 0.20 & 0.11 & 0.61 & 0.78 \\
        & $\{\text{EN}(A),\text{H}(A),\text{L}(A),r_d(A),r_p(A),r_s(A),\text{EA}(B),\text{L}(B)\} = 0.86$ & 0.20 & 0.12 & 0.62 & 0.77 \\[.5em]

        7 & $\{\text{EA}(A),\text{EN}(A),\text{H}(A),\text{L}(A),r_d(A),\text{EA}(B),\text{L}(B)\} = 0.85$ & 0.22 & 0.13 & 0.66 & 0.73 \\
        & $\{\text{EN}(A),\text{H}(A),\text{L}(A),r_d(A),r_p(A),\text{EA}(B),\text{L}(B)\} = 0.85$ & 0.20 & 0.12 & 0.62 & 0.77 \\
        & $\{\text{EA}(A),\text{EN}(A),\text{H}(A),\text{L}(A),r_d(A),r_p(A),r_s(A)\} = 0.84$ & 0.19 & 0.12 & 0.54 & 0.81 \\[.5em]

        6 & $\{\text{EA}(A),\text{EN}(A),\text{H}(A),\text{L}(A),r_d(A),r_p(A)\} = 0.83$ & 0.19 & 0.12 & 0.54 & 0.80 \\
        & $\{\text{EN}(A),\text{H}(A),\text{L}(A),r_d(A),\text{EA}(B),r_p(A)\} = 0.83$ & 0.19 & 0.12 & 0.54 & 0.80 \\
        & $\{\text{EN}(A),\text{H}(A),\text{L}(A),r_d(A),\text{EA}(B),\text{L}(B)\} = 0.83$ & 0.22 & 0.13 & 0.65 & 0.72 \\

        5 & $\{\text{EA}(A),\text{EN}(A),\text{H}(A),\text{L}(A),r_d(A)\} = 0.81$ & 0.19 & 0.12 & 0.57 & 0.79 \\
        & $\{\text{EN}(A),\text{H}(A),\text{L}(A),r_d(A),r_p(A)\} = 0.80$ & 0.19 & 0.12 & 0.55 & 0.81 \\
        & $\{\text{H}(A),\text{L}(A),r_d(A),\text{EA}(B),\text{L}(B)\} = 0.80$ & 0.22 & 0.14 & 0.66 & 0.71 \\[.5em]

        4 & $\{\text{EA}(A),r_s(A),r_d(B),r_p(B)\} = 0.77$ & 0.13 & 0.09 & 0.35 & 0.90 \\
        & $\{\text{EA}(A),\text{L}(A),r_p(A),r_s(A)\} = 0.77$ & 0.18 & 0.12 & 0.54 & 0.81 \\
        & $\{\text{EN}(A),\text{H}(A),\text{L}(A),r_d(A)\} = 0.77$ & 0.19 & 0.12 & 0.57 & 0.79 \\[.5em]

        3 & $\{\text{EA}(A),r_s(A),r_p(B)\} = 0.72$ & 0.13 & 0.09 & 0.35 & 0.90 \\
        & $\{\text{EA}(A),\text{L}(A),r_p(A)\} = 0.71$ & 0.18 & 0.12 & 0.54 & 0.81 \\
        & $\{\text{EA}(A),\text{L}(A),r_s(A)\} = 0.72$ & 0.19 & 0.12 & 0.57 & 0.80 \\[.5em]

        2 & $\{\text{EA}(A),r_s(A)\} = 0.69$ & 0.18 & 0.12 & 0.51 & 0.82 \\[.5em]

        1 & $\{r_s(A)\} = 0.34$ & 0.21 & 0.15 & 0.55 & 0.75 \\[.5em]

        \multicolumn{6}{l}{\itshape Monte Carlo dependency estimation (MCDE):} \\
        2 & $\{r_p(A),r_s(A)\} = 0.91$ & 0.20 & 0.14 & 0.53 & 0.77 \\[.5em]

        1 & $ \{r_s(A)\} = 0.89$ & 0.21 & 0.15 & 0.55 & 0.75 \\
        & $ \{r_p(A)\} = 0.89$ & 0.21 & 0.15 & 0.54 & 0.75 \\

        \midrule
        \multicolumn{2}{@{}l}{All 16 features (GBDT reference):} & 0.15 & 0.09 & 0.45 & 0.86 \\
        \bottomrule
    \end{tabular}
    \caption{Relevant feature subsets for the octet-binary compound semiconductors data set as found out by cumulative mutual information (CMI), multivariate maximal correlation analysis (MAC), universal dependency analysis (UDS), and Monte Carlo dependency estimation (MCDE). The table also shows statistics of constructed machine-learning models using the gradient boosting decision tree (GBDT) algorithm \citep{Friedman:2001} with 10-fold cross-validation: root-mean-squared error (RMSE), mean absolute error (MAE), maximum absolute error (MaxAE), and Pearson coefficient of determination ($r^2$). Units are in electronvolts (eV).}
    \label{tab:binary_octets_subsets_supplement}
\end{table*}

Formally, let $S_n$ denote the symmetric group of degree $n$, i.e., $S_n$ consists of all $n!$ bijections $\sigma: \{1\ldots n\} \to \{1\ldots n\}$. For a bijection $\sigma \in S_n$, we denote the permuted version of $Y$ as $Y_\sigma$. Then, for any cumulative contingency table $N \in \mathcal{M}(Y; Z)$ $S_n[N] = \{\sigma \in S_n: M(Y_\sigma; Z) = M \}$ denotes the permutations that result in $Z$. Let $\sigma \in S_n \setminus S_n[M]$. This means that $M_{ij}(Y; X) \neq M_{ij}(Y_\sigma; X)$ for at least one cell $i,j$. Further, denote the set of all indices of values of $X'$ that are projected down to $X$ by
\begin{equation}
    \pi^{-1}(j) = \{j': 1 \leq j' \leq C', \pi(X_{j'}') = X_j\} \ ,
\end{equation}
for which, by definition, follows that
\begin{equation}
    \sum_{j' \in \pi^{-1}(j)} M'_{ij'}(Y; X') \neq \sum_{j' \in \pi^{-1}(j)} M'_{ij'}(Y_\sigma; X') \ .
\end{equation}
Since for at least one index $j' \in \pi^{-1}(j)$ we get $M'_{ij'}(Y; X') \neq M'_{ij'}(Y_\sigma; X')$, we also find $\sigma \not\in S_n[M']$ and can conclude
\begin{equation}
    S_n[M] \supseteq \bigcup_{M' \in \pi^{-1}(M)} S_n[M'] \ .
    \label{eq:appendix_proof}
\end{equation}
Now let $N' \in \mathcal{M}(Y; X')$ with $\pi(N') \neq M$ and assume that $S_n[M] \supset S_n[M']$, i.e., there is a $\sigma \in S_n[M] \cap S_n[N']$. Let us denote $N = \pi(N')$. Since $S_n[M] \cap S_n[N] = \emptyset$, we know that $\sigma \not\in S_n[N]$. However, it follows from Eq.~\ref{eq:appendix_proof} that $\sigma \not\in S_n[N']$ -- a contradiction and, hence,
\begin{equation}
    S_n[M] = \bigcup_{M' \in \pi^{-1}(M)} S_n[M']
\end{equation}
and
\begin{equation}
    \begin{aligned}
        \mathcal P(Y; X \vert  M) & = \frac{\lvert S_n[M] \rvert}{\lvert S_n \rvert} = \sum_{M' \in \pi^{-1}(M)} \frac{\lvert S_n[M'] \rvert}{\lvert S_n \rvert} \\
        & = \sum_{M' \in \pi(M)} \mathcal P(Y; X \vert  M') \ .
    \end{aligned}
\end{equation}
\qed

\section{Gradient boosting decision trees}

We used LightGBM \citep{KeMengEtAl:2017}, a recent modification of the gradient-boosting decision trees algorithm \citep{Friedman:2001}. LightGBM improves the efficiency and scalability without sacrificing performance. The following settings were used and were found by hyper-parameter tuning: number of leaves (\verb|num_leaves|, 1\% of the number of samples), number of iterations (\verb|n_estimators|, 2000), and model depth (\verb|max_depth|, -1).

During the training, i.e., the model optimization, we performed a regularization to automatically select the inflection point at which the performance of the test data set begins to decrease while the performance of the training data set continues to improve. The data set was partitioned into 10 groups (so-called folds), using 9 folds to generate the model and the remaining fold to test the model (10-fold cross validation). To reduce variability, we performed five rounds of cross-validation with different partitions and averaged the rounds to obtain an estimate of the model's predictive performance. We monitored the $\ell_1$ and $\ell_2$ norms \citep{Friedman:2001,JamesWittenEtAl:2013} and simultaneously penalized the model optimization (``learning'') process on the 9 folds to minimize the squared residuals and the complexity of the model (\verb|eval_metric|, [``l1'', ``l2\_root'']), while stopping the learning process as soon as one metric of the remaining fold in the last $n = 50$ rounds did not improved (\verb|early_stopping_rounds|, 50).

\section{Feature-subset search}

We performed a feature-subset search using CMI, MAC, UDS, MCDE, and TCMI on the octet-binary compound semiconductors data set. Results of the feature-subset searches with TCMI can be found in Tab.~\ref{tab:binary_octets_subsets} and with CMI, MAC, UDS, and MCDE in Tab.~\ref{tab:binary_octets_subsets_supplement}. CMI, MAC, and MCDE dependence measures identify feature subsets with one atomic species only. Since the octet-binary compound semiconductor is uniquely determined by the atomic number of both atomic species, i.e., by at least one atomic property of each atomic species considered, CMI, MAC, and MCDE led to unreliable results. Due to issues with permutation and scale invariance (cf., Tab.~\ref{tab:scale_permutation_invariance}), UDS along with CMI, MAC, UDS, and MCDE were therefore not used further for model construction.

\small
\bibliographystyle{spphys}
\bibliography{bibliography}

\end{document}